\documentclass{article}

\usepackage{geometry}
\usepackage{hyperref}
\usepackage{xurl}
\hypersetup{colorlinks,citecolor=blue,urlcolor=blue, linkcolor=blue}
\usepackage{natbib}
\usepackage{amsmath, amsthm, amssymb}
\usepackage{caption}
\usepackage{subcaption}
\usepackage{float}
\usepackage{bbm}
\usepackage{algorithm}
\usepackage{algpseudocode}
\usepackage{pifont}
\usepackage[table,xcdraw]{xcolor}
\usepackage{graphicx}
\algrenewcommand\algorithmicrequire{\textbf{Input:}}
\algrenewcommand\algorithmicensure{\textbf{Output:}}

\usepackage{tikz}
\usepackage{tikzscale}
\usetikzlibrary{bayesnet}
\usetikzlibrary{positioning, calc}
\usetikzlibrary{decorations.pathmorphing}
\usetikzlibrary{arrows.meta}
\usetikzlibrary{shapes}

\usepackage{my_style}
\usepackage{comment}
\newcommand{\high}{\operatorname{high}}
\newcommand{\low}{\operatorname{low}}
\newcommand{\TS}{\operatorname{TS}}
\newcommand{\NA}{\operatorname{NA}}
\newcommand{\heart}{\operatorname{Heart}}
\newcommand{\sleep}{\operatorname{Sleep}}

\newcommand{\step}{\operatorname{Step}}
\newcommand{\sqrtstep}{\operatorname{SqrtStep}}
\newcommand{\weeklymood}{\operatorname{WeeklyMood}}
\newcommand{\burden}{\operatorname{Burden}}

\title{Dyadic Reinforcement Learning}

\author{Shuangning Li \thanks{University of Chicago Booth School of Business}
\and Lluís Salvat Niell \thanks{Departments of Physics and Mathematics, University of Manchester}
\and Sung Won Choi \thanks{Department of Pediatrics and Rogel Comprehensive Cancer Center, University of Michigan}
\and Inbal Nahum-Shani \thanks{Institute for Social Research, University of Michigan}
\and Guy Shani \thanks{Eli Broad College of Business, Michigan State University}
\and Susan A. Murphy \thanks{Susan A. Murphy holds concurrent appointments as a Professor of Statistics and Computer Science at Harvard University and as an Amazon Scholar. This paper describes work performed at Harvard University and is not associated with Amazon.}
}

\begin{document}

\maketitle

\begin{abstract}
Mobile health aims to enhance health outcomes by delivering interventions to individuals as they go about their daily life. The involvement of care partners and social support networks often proves crucial in helping individuals managing burdensome medical conditions. This presents opportunities in mobile health to design interventions that target the dyadic relationship---the relationship between a target person and their care partner---with the aim of enhancing social support. In this paper, we develop dyadic RL, an online reinforcement learning algorithm designed to personalize intervention delivery based on contextual factors and past responses of a target person and their care partner. Here, multiple sets of interventions impact the dyad across multiple time intervals. The developed dyadic RL is Bayesian and hierarchical. We formally introduce the problem setup, develop dyadic RL and establish a regret bound. We demonstrate dyadic RL's empirical performance through simulation studies on both toy scenarios and on a realistic test bed constructed from data collected in a mobile health study.
\end{abstract}

\section{Introduction}
\label{section:introduction}
Mobile health studies involve delivering digital interventions to individuals with the aim of promoting healthy behaviors and improving health outcomes. However, health management often requires the involvement of more than just the individual. For example, family care partners play an essential role in managing intense and high-risk cancer procedures \citep{psihogios2019preferences, psihogios2020adherence}. For adolescents and young adults with cancer, up to 73\% of family care partners are primarily responsible for managing cancer-related medications \citep{psihogios2020adherence}. Similar to cancer treatment, social support networks are also crucial for individuals with addiction or mental illness \citep{mitchell1980task, siegel1994strengthening,panebianco2016personal}. 
In all the above examples, the care partner plays a critical role in the health management of the target person (e.g., young adult with cancer). The pair of the target person and the care partner is known as a \textit{dyad}, and the relationship between them is known as a \textit{dyadic relationship}. In the context of mobile health studies, this presents an opportunity to harness the power of the dyad and develop interventions that target the dyadic relationship as well as both members of the dyad \citep{carlozzi2021app, koblick2023pilot}. The goal of such interventions is to enhance motivation and social support for the target person, which can lead to improved health outcomes.

\begin{figure}
    \centering
    \includegraphics[width = 0.55\textwidth]{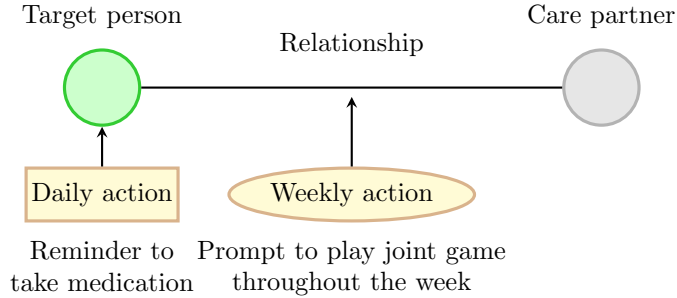}
    \caption{Simplified illustration of the dyad. Daily actions are directed towards the target person, while weekly actions aim to enhance the relationship between the target person and their care partner.}
    \label{fig:social_network}
\end{figure}

 At present, we are involved in the early stages of designing a reinforcement learning (RL) algorithm that will be deployed in the \textbf{\textit{ADAPTS HCT}} clinical trial. ADAPTS HCT aims to enhance medication adherence in adolescents with allogeneic hematopoietic stem cell transplantation (HCT) over the 14 weeks following transplantation. The developed RL algorithm will concern multiple types of sequential decisions, including whether to send motivational prompts to the adolescents (targeting the adolescent), and whether to send prompts to both the adolescents and their parents to encourage them to play a joint game (targeting the dyadic relationship). The game is designed to foster social support and collaboration. See Figure~\ref{fig:social_network} for a simplified illustration of the dyad and the interventions (actions).  

To make first steps in constructing the RL algorithm, we focus on settings where different sets of actions are appropriate for use at varying time intervals. For example, in ADAPTS HCT, daily motivational prompts could be sent to adolescents to improve medication adherence on that particular day, whereas the weekly collaborative game in ADAPTS HCT lasts throughout the week. Similarly, a catalog company, might use monthly actions for campaigns (e.g., discounts offered over a month), while actions on a smaller time scale might highlight different types of products.

We face a number of key challenges when dealing with this type of problem. These challenges include:
\begin{enumerate}
\item Impact over different time scales: Different sets of actions impact the environment (individuals within a dyad) over different time scales.
\item High noise levels: There is a substantial amount of noise in both state transitions and rewards.
\item Need for interpretability: Our results and methodologies need to be interpretable to health scientists to allow critical feedback and ensure applicability. 
\end{enumerate}

The dyadic RL algorithm, developed here, is tailored to explicitly accommodate the aforementioned challenges. This algorithm interacts with episodes each of finite length (in the context of ADAPTS HCT, each episode involves interacting with a dyad). To insure robustness of the dyadic RL algorithm against a further challenge that is likely to occur in mobile health, that is, the dyads are likely heterogeneous, we evaluate dyadic RL in a testbed environment that incorporates this heterogeneity (More details can be found in Section \ref{section:application}).

The developed dyadic RL algorithm is hierarchical with two levels. The lower level involves a finite number of time periods within a time block; the algorithm learns policies in this contextual finite-horizon decision-making setting \citep{benjamins2021carl,benjamins2022contextualize}. Conversely, at the higher level, which involves time blocks within an episode, the algorithm focuses on learning a contextual bandit policy. This hierarchical arrangement not only promotes learning across different episodes but also facilitates learning within episodes, thereby accelerating the overall learning process.

The novelties presented in our paper are threefold. 

\begin{enumerate}
    \item Firstly, we utilize domain-specific knowledge, particularly insights about the duration of effects of various actions, to develop a hierarchical algorithm. 
    This hierarchical structure enables learning not only between episodes (e.g., dyads) but also within episodes,
    significantly enhancing the speed of learning.
    Further, we include the high-level states and actions into the low-level states. This is crucial because the high-level actions' impacts affect states over extended periods. This step transforms the non-Markovian environment back into a Markovian one.
    
    \item Secondly, we devise a novel reward construction for the high-level agent within the hierarchical RL algorithm. This reward construction uses an estimate of the $Q$-function, which aids in reducing noise in the original reward.
    
    \item Thirdly, we present a regret bound for the proposed hierarchical RL algorithm within a tabular setting. Although hierarchical algorithms, as discussed further in Section \ref{section:related_work}, often do not attain the standard regret rate, we prove that the dyadic RL algorithm does. In particular, we attain a regret bound of $\sqrt{K W}$, where $K$ represents the number of episodes and $W$ denotes the number of time blocks in each episode. 

\end{enumerate}

The rest of the paper is organized as follows:
In Section~\ref{section:dyadic_rl}, we provide a formal introduction to the problem set up and present the dyadic RL algorithm.
In Section~\ref{section:theory}, we establish the regret bound in a tabular setting.
We demonstrate the empirical performance of the algorithm through simulation studies on toy scenarios in Section~\ref{section:simulations}. Finally, in Section~\ref{section:application}, we test the proposed algorithm on a realistic test bed constructed from data collected in a previous mobile health study. 

\subsection{Related work}
\label{section:related_work}

\paragraph{Hierarchical Reinforcement Learning}
The hierarchical reinforcement learning family of methods addresses complex problems 
through decomposing these problems into more manageable subtasks. The main advantages of hierarchical RL include increased learning speed \citep{dayan1992feudal}, incorporation of transfer learning \citep{mehta2008transfer}, promotion of generalization \citep{sohn2018hierarchical}, and enhanced exploration \citep{dietterich2000hierarchical}. \citet{wen2020efficiency} present a theoretical framework for evaluating the statistical and computational efficiency of hierarchical RL methods. They demonstrate that the expected regret can be diminished when both the environment and the algorithm display a hierarchical structure. Unlike most hierarchical RL algorithms, which use the options framework \citep{sutton1999between}, our method does not plan over extended time horizons at the highest hierarchy. 

The first and most extensive class of hierarchical RL problems is characterized by high-dimensional or continuous state spaces. Here, hierarchy relies on state abstraction \citep{andre2002state, li2006towards}, temporal abstraction \citep{sutton1999between} or hierarchical abstraction \citep{dayan1992feudal, parr1997reinforcement, dietterich1998maxq}, facilitating more efficient computation. Despite achieving faster learning speeds, the agents might exhibit sub-optimal behavior \citep{dayan1992feudal}. These methods may yield a \textit{hierarchically} optimal policy \citep{parr1997reinforcement} or a \textit{recursively} optimal policy \citep{kaelbling1993hierarchical}, but not a \textit{globally} optimal policy.

The second class of methods also considers the reduction of dimensionality in large state spaces. However, unlike the first class, the agents in this category converge to the globally optimal policy. \citet{infante2022globally} applies the hierarchical RL formalism, as introduced by \citet{wen2020efficiency}, to a unique kind of MDPs, specifically, the so-called linearly-solvable MDPs. By representing value functions on multiple levels of abstraction and using the compositionality of subtasks to estimate the optimal values of the states in each partition, \citet{infante2022globally} propose an approach capable of learning the globally optimal policy.

Our approach distinguishes itself from the first two classes of methods in several key aspects: While hierarchical structures accelerate learning for both our method and the first class of methods, our employment of hierarchical methods is primarily to manage noise and to ensure interpretability, while the first class utilizes hierarchical RL to accommodate high dimensional state spaces. In relation to the second class of methods, the high noise combined with likely small effects of the high-level actions on the subsequent time block's states motivates our use of a bandit structure at the high level of hierarchy. If there are truly no effects of the actions on the subsequent time block's states, then our method will converge to the optimal policy. In the presence of delayed effects across time blocks, the simulation studies (as detailed in Sections~\ref{section:simulations} and \ref{section:application}) illustrate that our algorithm continues to perform well.

\paragraph{Reinforcement Learning on Social Networks} RL has been established as a practical framework for decision-making in single large-scale social networks. Much of the existing literature in this field centers on influence maximization, the goal of which is to optimally identify nodes that exert the maximum influence on either known \citep{goindani2020social, chen2021contingency} or unknown social networks \citep{kamarthi2019influence, ali2020addressing, li2021claim}. The methodology developed for influence maximization has found many applications, such as the optimization of behavioral interventions for disease prevention \citep{yadav2015preventing, yadav2016using, yadav2019influence}. Our approach in this paper differs from the above literature in a few ways. (1) Instead of focusing on a single large-scale network, we consider episodes of smaller networks. (2) Our RL algorithm concerns actions on both nodes (people) and edges (relationships) within the network. (3) Unlike the aforementioned literature, different treatment actions may be delivered over different time scales with associated effects over the different time scales. 

\paragraph{Reinforcement Learning in Mobile Health Studies} 
In mobile health studies, RL algorithms have successfully assisted with personalizing treatment \citep{tomkins2020rapidly, gonul2021reinforcement, ghosh2023did}, and addressed challenges due to noisy data and user burden \citep{liao2020personalized}. These algorithms can be used for a range of mobile health goals, such as improving medication adherence \citep{etminani2021improving}, promoting weight loss \citep{forman2019can}, encouraging physical activity \citep{yom2017encouraging, zhou2018personalizing, liao2020personalized}, and facilitating healthy habit formation \citep{hoey2010automated, trella2022reward}. To the best of our knowledge, once fully developed, our dyadic RL algorithm will be the first RL algorithm used in a mobile health study that (1) employs a hierarchical structure, and (2) administers interventions to not only impact the target person, but also their relationship with their care partner.


\section{Dyadic RL}
\label{section:dyadic_rl}

\subsection{Notation and setup}
Assume that dyads join the trial sequentially. Let $k = 1, \dots, K$ represent the dyad number. In each dyad's trial, there exists a sequence of time blocks, each with a length of $H$. We use $w = 1, 2, \dots, W$ to denote the time block number, and $h = 1, 2, \dots, H$ for the time period number within each time block. In the context of ADAPTS HCT, a time block equates to a week, while a time period corresponds to a day. Thus in ADAPTS HCT, $H = 7$ days and $W=14$ weeks.

For dyad $k$, each time block $w$ has a high-level state $s^{\high}_{k,w} \in \mathcal{S}^{\high}$. Within that time block $w$, each time period $h$ has a low-level state $s^{\low}_{k, w,h} \in \mathcal{S}^{\low}$. In the context of ADAPTS HCT, high-level states could comprise weekly measurements of the quality of the dyadic relationship, and low-level states could comprise various daily measurements related to patients and their parents (care partners), such as their daily step count or amount of sleep. The low-level reward is denoted by $r^{\low}_{k,w,h} \in \mathbb{R}$ at each time period $h$. This reward can be based on the degree to which  the patient adhered to their medications  in ADAPTS HCT.

There are two types of actions: the high-level action and the low-level action. The high-level action, $a_{k,w}^{\high} \in \mathcal{A}^{\high}$, occurs at the start of each time block $w$. In contrast, the low-level action, $a^{\low}_{k,w,h} \in \mathcal{A}^{\low}$, takes place at the beginning of time period $h$ within the time block $w$. In ADAPTS HCT, these two types of actions correspond to two types of interventions: one targeting the patient and the other targeting the dyadic relationship. The low-level action involves sending engagement messages to patients to help them maintain their motivation to take their medication. However, the high-level action consists of encouraging both the patient and their care partners to engage in a joint activity, such as a game."

To summarize the notation, if we have an algorithm that chooses actions, we will observe the following trajectory of data for dyad $k$:
\[
\dots, \underbrace{\, s_{k,w}^{\high}, a_{k,w}^{\high}, \quad s^{\low}_{k,w,1}, a^{\low}_{k,w,1}, r^{\low}_{k,w,1}, \, s^{\low}_{k,w,2}, a^{\low}_{k,w,2}, r^{\low}_{k,w,2}, \,\dots,\, s^{\low}_{k,w,H}, a^{\low}_{k,w,H}, r^{\low}_{k,w,H}}_{\text{time block } w }, \quad \underbrace{s_{k,w+1}^{\high}, a_{k,w+1}^{\high}, \,\dots\,}_{\text{time block } w+1 }, \,\dots
\]

\begin{figure}[t]
    \centering
    \includegraphics[width = \textwidth]{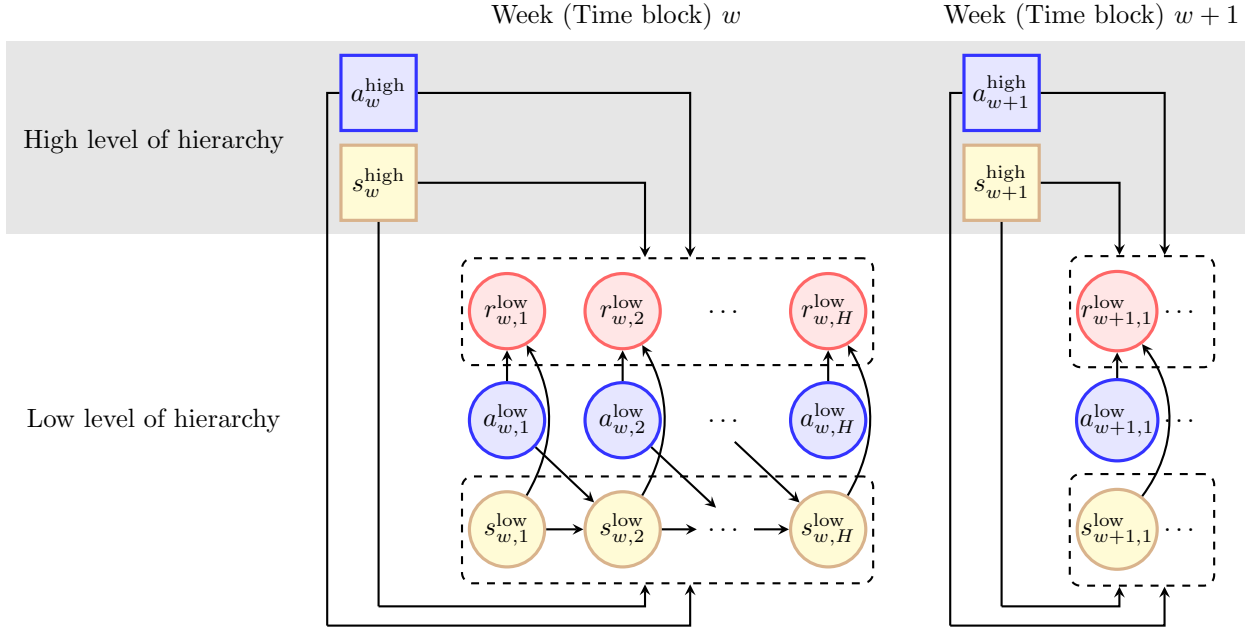}
    \caption{A directed acyclic graph illustrating the relationship of the variables in episode $k$. Here, $s^{\low}_{k,w,h}$, $a^{\low}_{k,w,h}$ and $r^{\low}_{k,w,h}$ represent the low-level state, action and reward at time period $h$ of time block $w$ respectively. Meanwhile, $s^{\high}_{k,w}$ and $a^{\high}_{k,w}$ represent the high-level state and action in time block $w$.}
    \label{fig:weekly_DAG}
\end{figure}

In Figure \ref{fig:weekly_DAG}, we provide an illustration that depicts the relationship between the variables. This relationship is based on Approximations \ref{appr:episodic_mdp} and \ref{appr:hier_structure}, which are introduced in the following Section \ref{subsection:appr_envi}.

Finally, we introduce the notation for feature mappings. A feature mapping is a function that maps a data vector to the feature space $\mathbb{R}^p$ for some dimension $p$. In the following sections, we use $\phi$ or $\psi$ to denote different feature mappings. 

\subsection{Approximations for the environment}
\label{subsection:appr_envi}
We would like to design a reinforcement learning algorithm that chooses the actions which optimize the cumulative reward. To this end, we make a few approximations for the environment, each of which are made in line with insights derived from domain science related to ADAPTS HCT.

We start with defining the history prior to dyad $k$, time block $w$, and time period $h$ as: 
\begin{equation}
\begin{split}
\mathcal{H}_{k,w,h} =
&\Big\{\p{s^{\low}_{\breve{k}, \breve{w}, \breve{h}}, a^{\low}_{\breve{k}, \breve{w}, \breve{h}}, r^{\low}_{\breve{k}, \breve{w}, \breve{h}}}:  \breve{k} < k \text{ or } (\breve{k} = k \text{ and } \breve{w}<w) \text{ or } (\breve{k} = k, \breve{w}=w \text{ and } \breve{h}<h) \Big\} \cup\\
&\qquad \qquad\qquad\qquad\qquad\Big\{\p{s^{\high}_{\breve{k}, \breve{w}}, a^{\high}_{\breve{k}, \breve{w}}}:  \breve{k} < k \text{ or } (\breve{k} = k \text{ and } \breve{w}\leq w)  \Big\}. 
\end{split}
\end{equation}
Note that $\mathcal{H}_{k,w,h}$ includes the high-level state and action in time block $w$. Similarly, we define the history prior to episode $k$, time block $w$ as:
\begin{equation}
\begin{split}
\mathcal{H}_{k,w,0} =
&\Big\{\p{s^{\low}_{\breve{k}, \breve{w}, \breve{h}}, a^{\low}_{\breve{k}, \breve{w}, \breve{h}}, r^{\low}_{\breve{k}, \breve{w}, \breve{h}}}:  \breve{k} < k \text{ or } (\breve{k} = k \text{ and } \breve{w}<w) \Big\} \cup\\
&\qquad \qquad\qquad\qquad\qquad\Big\{\p{s^{\high}_{\breve{k}, \breve{w}}, a^{\high}_{\breve{k}, \breve{w}}}:  \breve{k} < k \text{ or } (\breve{k} = k \text{ and } \breve{w}< w)  \Big\}. 
\end{split}
\end{equation}
Note, $\mathcal{H}_{k,w,0}$ does not include the high-level state and action in time block $w$.

\begin{appr}[Episodic MDP]
\label{appr:episodic_mdp}
For each $w \in \cb{1,\dots, W}, h \in \cb{0,\dots, H-1}$, there exists a Markov transition kernel $P_{w,h}$ such that
\begin{equation}
\begin{split}
P_{w,h}\p{s; s^{\high}_{k, w}, a^{\high}_{k,w}, s^{\low}_{k, w, h}, a^{\low}_{k,w, h}}
&=\PP{s^{\low}_{k, w, h+1}=s \mid \mathcal{H}_{k,w,h+1}}, \textnormal{ if } h \geq 1, \\
P_{w,0}\p{s; s^{\high}_{k, w}, a^{\high}_{k,w}}
&=\PP{s^{\low}_{k, w, 1}=s \mid \mathcal{H}_{k,w,1}} \textnormal{ if } h = 0,
\end{split}
\end{equation}
for any $k \in \cb{1,\dots, K}$. 
Furthermore, for each $w \in \cb{1, \dots, W}$, there exists a Markov transition kernel $P^{\high}_{w}$ such that 
\begin{equation}
P^{\high}_{w}\p{s; s^{\high}_{k, w}, a^{\high}_{k,w}, s^{\low}_{k, w, H}, a^{\low}_{k,w, H}}
=\PP{s^{\high}_{k, w+1}=s \mid \mathcal{H}_{k,w+1,0}}. 
\end{equation}

In addition, for each $w \in \cb{1,\dots, W}, h \in \cb{1,\dots, H}$, there exists a reward function $R_{w,h}$ such that
\begin{equation}
R_{w,h}\p{r; s^{\high}_{k, w}, a^{\high}_{k,w}, s^{\low}_{k, w, h}, a^{\low}_{k,w,h}}
=\EE{r^{\low}_{k, w, h}\mid s^{\low}_{k, w, h}, a^{\low}_{k,w, h}, \mathcal{H}_{k,w,h}},
\end{equation}
for any $k \in \cb{1,\dots, W}$. 
\end{appr}

Approximation \ref{appr:episodic_mdp} carries several implications. Firstly, Approximation \ref{appr:episodic_mdp} posits that the environment is episodic, with each dyad representing an episode. From now on, when we refer to the $k$-th dyad, we may directly call it the $k$-th episode. This approximation specifically implies that actions targeting one dyad won’t affect other dyads. Moreover, the state transition functions and reward distributions are the same across all dyads. In the context of ADAPTS HCT, each dyad joins the trial seperately, making it plausible to assume there is no spillover effect from one dyad to another. While there may be heterogeneity across dyads in ADAPTS HCT, assuming homogeneity enables the pooling of data across them. Such pooling reduces noise and is especially beneficial in mobile health studies. In these studies, both state transitions and reward distributions often contain high levels of noise. Managing this noise becomes a crucial objective when developing RL algorithms.

Secondly, Approximation \ref{appr:episodic_mdp} indicates that the environment assumes a Markovian structure when both the high-level state and action are incorporated into the states. The inclusion of high-level states and actions is crucial because the high-level action can influence the environment throughout an entire time block. In the ADAPTS HCT setting, the high-level action is a motivational prompt sent at the beginning of the week, encouraging the patient and care partner dyad to participate in a joint game. Since this game might span the whole week, the effects of the weekly prompt are not confined to just the immediate day, indicating a prolonged impact. This characteristic underscores a non-Markovian structure without state reconstruction, adding a distinctive layer of complexity to our problem.

\begin{appr}[Hierarchical Structure]
\label{appr:hier_structure}
For every $w$ and $h$, $P_{w,h} = P_{1,h}$ and $\mathcal{R}_{w,h} = R_{1,h}$,  where $P_{w,h}$ and $R_{w,h}$ are defined in Approximation \ref{appr:episodic_mdp}. Furthermore, there exist functions $\nu_w$ such that
\begin{equation}
\nu_w\p{s; s^{\high}_{k, w}} = \PP{s^{\high}_{k, w+1}=s \mid \mathcal{H}_{k,w+1,0}}. 
\end{equation}
\end{appr}

Approximation \ref{appr:hier_structure} also carries several implications. Firstly, it suggests that the environment operates as a contextual bandit environment at the level of time blocks, meaning actions within one time block don't affect subsequent time blocks. Secondly, Approximation \ref{appr:hier_structure} implies homogeneity across time blocks. Specifically, it indicates that the state transition function and the reward function, when conditioned on the high-level state and high-level action, remain consistent across all time blocks.

The plausibility of this approximation in mobile health studies is due to the noise level. Over the course of a time block in mobile health studies, significant noise can accumulate, resulting in a low signal-to-noise ratio and thus posing significant challenges in accurately modeling and capturing delayed effects that span multiple time blocks. To address this, we make Approximation \ref{appr:hier_structure}  that allows us to consider simpler finite horizon algorithms. Importantly, our objective here is to develop an algorithm that will select actions that result in high rewards in spite of the delayed effects in this noisy environment. In Sections \ref{section:simulations} and \ref{section:application}, we present empirical evidence demonstrating that, even when facing long-term delayed effects, our proposed algorithm---which implements a contextual bandit algorithm at the higher level of hierarchy---maintains desirable performance.

Lastly, we  explain why we refer to Approximation \ref{appr:hier_structure} as a “hierarchical structure” approximation. Fundamentally, each time block can be viewed as a subtask, and the knowledge from previous subtasks can be used for the current ones. In connection with the hierarchical RL literature, the high-level and low-level states and actions align with the ideas of task-independent and task-related spaces presented in \citep{konidaris2006autonomous, konidaris2007building, florensa2017stochastic, gupta2017learning, li2019hierarchical}. Each time block can be thought of as a contextual MDP, as described in \citep{hallak2015contextual, benjamins2021carl, benjamins2022contextualize}: the values of the high-level state and action set the context, which in turn modifies the transition functions of the MDP. Hence, each time block, with identical values of the high-level state and action, is identified as a unique task for the agent to handle. From this perspective, the high-level state and action are considered task-related variables, while the low-level state embodies the task-independent component.  
In Appendix \ref{appendix:wen}, we provide an alternative interpretation of the hierarchical structure, employing the framework proposed by \citet{wen2020efficiency}.

\subsection{Dyadic RL algorithm}
\label{subsection:algorithm}

Building on the approximations for the environment, we develop a hierarchical RL algorithm tailored to facilitate rapid learning within an episode. As detailed in Section \ref{section:related_work}, hierarchical RL algorithms simplify complex problems by breaking them down into manageable subtasks, grouping together those subtasks with shared characteristics. In line with our analysis in Section~\ref{subsection:appr_envi}, when developing our algorithm, we take advantage of the inherent hierarchical structure of our problem by treating each time block as a separate subtask. Specifically, our algorithm operates on two hierarchical levels: a high-level agent responsible for selecting high-level actions at the beginning of each time block, and a low-level agent tasked with choosing low-level actions for each time period within a time block. 

\begin{algorithm}
\caption{Randomized Least-Squares Value Iteration}
\label{alg:rlsvi_generic}
\begin{algorithmic}
\Require 
Horizon $H$, Data $\cb{s_{\breve{w},1}, a_{\breve{w},1}, r_{\breve{w},1}, \dots, s_{\breve{w},H}, a_{\breve{w},H}, r_{\breve{w},H}}_{\breve{w} = 1\dots, w-1}$;
Feature mappings $\phi_1, \dots, \phi_H$; Parameters $\lambda, \sigma>0$.
\For{$h = H, \dots, 1$}
\begin{enumerate}
\item Generate regression problem $X \in \mathbb{R}^{(w-1) \times p}, y \in \mathbb{R}^{w-1}$ :
\[\begin{array}{l}
X = \left[\begin{array}{c}
\phi_h\left(s_{1,h}, a_{1,h} \right) ^\top \\
\vdots \\
\phi_h\left(s_{w-1, h}, a_{w-1, h}\right) ^\top
\end{array}\right], \\
\quad \\
y_{\breve{w}} = \left\{\begin{array}{ll}
r_{\breve{w}, h}+\max _\alpha \tilde{\theta}_{w, h+1}^\top \phi_{h+1}\left(s_{\breve{w}, h+1}, \alpha\right) & \text { if } h < H, \\
r_{\breve{w}, h} & \text { if } h = H.
\end{array}\right.
\end{array}\]

\item Bayesian linear regression for the value function:
\[\begin{array}{l}
\bar{\theta}_{w,h} \leftarrow \frac{1}{\sigma^2}\left(\frac{1}{\sigma^2} X^{\top} X + \lambda I\right)^{-1} X^{\top} y, \\
\Sigma_{w,h} \leftarrow\left(\frac{1}{\sigma^2} X^{\top} X+\lambda I\right)^{-1}.
\end{array}\]

\item Sample $\tilde{\theta}_{w,h} \sim N\left(\bar{\theta}_{w,h}, \Sigma_{w,h}\right)$ from posterior.
\end{enumerate}
\EndFor
\Ensure $\tilde{\theta}_{w, 1}, \dots, \tilde{\theta}_{w, H}$
\end{algorithmic}
\end{algorithm}

At the low level of hierarchy, Approximation~\ref{appr:hier_structure} suggests using a finite-horizon RL algorithm with the horizon set to $H$, which corresponds to the length of a time block. This is due to the fact that time blocks are independent and exhibit similar structures. We particularly focus on the randomized least-squares value iteration (RLSVI) algorithm \citep{osband2016generalization}, as outlined in Algorithm \ref{alg:rlsvi_generic}. Recall that in a finite horizon problem, there is a $Q$-function for each time period. The $Q$-function at time $h$ is the expected sum of rewards, until the end of the horizon $H$, starting at time $h$ given (state, action) at time $h$. The Q-function depends on the policies used at times $h+1, h+2, \dots, H$. In Algorithm \ref{alg:rlsvi_generic}, we use a linear model for the time $h$ $Q$-function in terms of a feature vector $\phi_h(s,a)$: 
\begin{equation}
    Q_{h,\theta}(s,a) = \theta^{\top}_h \phi_h(s,a). 
\end{equation}
For every time period within a time block, the algorithm produces a random draw $\tilde{\theta}$ from the posterior distribution of parameter $\theta$. The construction of this posterior distribution depends on the reward of the current time period and the estimates of the $Q$-function for the subsequent time period. 

Meanwhile, we note that the high-level state and high-level action can exert delayed impacts throughout the time block. This highlights the need to restructure the state when implementing RLSVI at the low-level of hierarchy. Specifically, as discussed after Approximation \ref{appr:episodic_mdp}, we incorporate the high-level state and high-level action into the low-level state. 

It is worth pointing out that the original intention behind the design of the RLSVI algorithm was to enable efficient exploration. In our case, we have also selected this algorithm due to its Bayesian characteristics. First, the RLSVI algorithm utilizes straightforward linear regression, significantly enhancing its interpretability. Second, mobile health trials often operate with small sample sizes. In such conditions, the strength of Bayesian methods, such as RLSVI, becomes evident: Bayesian methods regularize learning, which improves estimation error when there is limited data \citep{jackman2009bayesian}.

At the high level of hierarchy, Approximation \ref{appr:hier_structure} directs our attention towards implementing a contextual bandit algorithm at the level of time blocks. Specifically, we consider the Thompson Sampling algorithm \citep{russo2018tutorial}, which can be viewed as a special case of RLSVI when the horizon $H = 1$.
Now, a critical question arises: what should this high-level agent consider as the reward? Ideally, the reward should be a denoised version of the sum of rewards in the coming time block, presuming that the low-level agent performs optimally. However, in reality, the low-level agent, during the learning process, does not possess complete knowledge of the optimal policy, hence, we can only estimate the aforementioned quantity. Thus at each time block $w$, we reconstruct all rewards at prior time blocks $\breve{w}\le w$: 
\begin{equation}
\label{eqn:reward_const}
\tilde{r}^{\high}_{\breve{w}} = \max_{\alpha} \tilde{\theta}_{w, 1} ^\top \phi_1 \left(s^{\high}_{\breve{w}}, a^{\high}_{\breve{w}}, s^{\low}_{\breve{w}, 1}, \alpha \right).
\end{equation}
Here, $\phi_1$ is the time period $h=1$ feature mapping of state values. The term $\tilde{\theta}_{w, 1}$ represents a draw from the most recent posterior distribution of the parameter $\theta_1$ within the RLSVI considered by the low-level agent. The inner product, $\tilde{\theta}_{w, 1} ^\top \phi_1 \left(s^{\high}_{\breve{w}}, a^{\high}_{\breve{w}}, s^{\low}_{\breve{w}, 1}, \alpha \right)$, provides an estimation of the $Q$-function on the first time period of time block $\breve{w}$. By maximizing over the action space, we obtain an estimate of the expected sum of rewards for time block $\breve{w}$, presuming optimal performance by the low-level agent.

To the best of our knowledge, this reward construction is the first in the hierarchical RL literature that constructs rewards for the high-level agent. Previously in the hierarchical RL literature, reward construction has primarily focused on low-level agents: the high-level agent receives the primary reward, while the low-level agent works with a subtask reward. This subtask reward is usually assigned by the high-level agent, and it can either be handcrafted or learned during the RL algorithm implementation \citep[][among others]{kulkarni2016hierarchical, vezhnevets2017feudal, nachum2018data, li2019hierarchical}.
This reward construction also connects to the concept of bootstrapping in the classical RL literature. Rather than directly using the observed reward, we update our choice of parameter based on prior estimates of parameters. Bootstrapping is known to be helpful in reducing variance \citep{sutton2018reinforcement}.

Incorporating all the above elements, we present our dyadic RL algorithm in Algorithm~\ref{alg:rlsvi_greedy_more_episodes}.

\begin{algorithm}
\caption{Dyadic RL}
\label{alg:rlsvi_greedy_more_episodes}
\begin{algorithmic}
\Require Feature mappings for the algorithm at the low level of hierarchy $\phi_1, \dots, \phi_{H}$; Feature mapping for the algorithm at the high level of hierarchy $\psi$; Parameters $\cb{\sigma_{k,w}, \lambda_{k,w}, \sigma_{\TS,k,w}, \lambda_{\TS,k,w}}_{k,w > 0}$. 
\State $\text{Data}^{\high} = \cb{}$ and $\text{Data}^{\low} = \cb{}.$
\For{Episode $k = 1, 2, \dots, K$}
\For{Time block $w = 1, 2, \dots, W$}
\State Observe $s_{k,w}^{\high}$.
\State Input $H = 1$, $\text{Data}^{\high}$ and $\sigma_{k,w,\TS}, \lambda_{k,w,\TS}$ into Algorithm 
\ref{alg:rlsvi_generic} and get output $\tilde{\beta}_{k,w}$. 
\State Sample $a^{\high}_{k,w} \in \operatorname{argmax}_{\alpha \in \mathcal{A}}\tilde{\beta}_{k,w} ^\top \psi \left(s^{\high}_{k,w}, \alpha \right)$.
\State Input $H$, $\text{Data}^{\low}$ and $\sigma_{k,w}, \lambda_{k,w}$ into Algorithm 
\ref{alg:rlsvi_generic} and get output $\tilde{\theta}_{k,w,1}, \dots, \tilde{\theta}_{k,w, H}$. 
\For{Time period $h = 1, \dots, H$}
\State Observe $s_{k,w,h}^{\low}$.
\State Sample $a_{k,w, h}^{\low} \in \operatorname{argmax}_{\alpha} \tilde{\theta}_{k,w, h} ^\top \phi_h \left(s^{\high}_{k,w}, a^{\high}_{k,w}, s^{\low}_{k,w, h}, \alpha \right)$.
\State Observe $r_{k,w,h}^{\low}$.
\State Add $\p{\p{s_{k,w}^{\high}, a_{k,w}^{\high},s_{k,w, h}^{\low}}, a_{k,w,h}^{\low}, r_{k,w,h}^{\low}}$ to $\text{Data}^{\low}$.
\EndFor
\State Set $\tilde{r}^{\high}_{\breve{k},\breve{w}} = \max_{\alpha} \tilde{\theta}_{k, w, 1} ^\top \phi_1 \left(s^{\high}_{\breve{k},\breve{w}}, a^{\high}_{\breve{k},\breve{w}}, s^{\low}_{\breve{k}, \breve{w}, 1}, \alpha \right)$ for all prior $(\breve{k},\breve{w})$, i.e., $(\breve{k},\breve{w})$ such that $\breve{k} \leq k$ or ($\breve{k} = k$ and $\breve{w} \leq w$), update all $\tilde{r}^{\high}_{\breve{k},\breve{w}}$ in $\text{Data}^{\high}$. 
\State Add $\p{s_{k,w}^{\high}, a_{k,w}^{\high},\tilde{r}_{k,w}^{\high}}$ to $\text{Data}^{\high}$.
\EndFor
\EndFor
\end{algorithmic}
\end{algorithm}

We would like to underscore once more that through the application of hierarchical RL, we gain the ability to learn both within and across episodes. In the absence of this hierarchical structure, one can primarily learn across episodes, but with limited capacity for within-episode learning.

\section{A Regret Bound}
\label{section:theory}

In this section, we establish a regret bound of our dyadic RL algorithm in a tabular setting. Throughout the section, we operate under the assumption that Approximations \ref{appr:episodic_mdp} and \ref{appr:hier_structure} hold true. The proof derives from that of the non-hierarchical regret bound in \citep{russo2019worst}. We begin by introducing notations and definitions, then transition into stating the primary regret bound.

We introduce notation following \citep{russo2019worst}. Let 
\begin{equation}
R_{h}(s^{\high}, a^{\high}, s^{\low},  a^{\low}) = \EE{r^{\low}_{k,w,h} \mid s^{\high}_{k, w} = s^{\high}, a_{k,w}^{\high}= a^{\high},
s^{\low}_{k, w, h} = s^{\low}, a_{k, w, h}^{\low} = a^{\low}}.
\end{equation}
Here, thanks to Approximation \ref{appr:hier_structure}, we are able to drop the subscript $k$ and $w$ in $R$. 
A deterministic Markov policy $\pi=\left(\pi_0, \pi_1, \ldots, \pi_H\right)$ is a sequence of functions, where $\pi_0: \mathcal{S}^{\high} \rightarrow \mathcal{A}^{\high}$ prescribes a high-level action and $\pi_h: \mathcal{S}^{\high} \times \mathcal{A}^{\high} \times \mathcal{S}^{\low} \rightarrow \mathcal{A}^{\low}$ prescribes a low-level action for $h \geq 1$. We let $\Pi$ denote the space of all such policies. For $h \geq 1$, we use $V_h^\pi \in \mathbb{R}^{\abs{\mathcal{S}^{\high}}\abs{\mathcal{A}^{\high}}\abs{\mathcal{S}^{\low}}}$ to denote the value function associated with policy $\pi$ in the sub-episode consisting of periods $\{h, \ldots, H\}$. We use $V_0^\pi \in \mathbb{R}^{\abs{\mathcal{S}^{\high}}}$ to denote the value function associated with policy $\pi$ at the beginning of the time block after observing the high-level state. 
More specifically, for $h \geq 1$, 
\begin{equation}
V_h^\pi(s^{\high}, a^{\high}, s^{\low})
= \EE[\pi]{\sum_{\breve{h} = h}^H r^{\low}_{k,w,\breve{h}} \mid s^{\high}_{k, w} = s^{\high}, a_{k,w}^{\high}= a^{\high},
s^{\low}_{k, w, \breve{h}} = s^{\low}},
\end{equation}
and 
\begin{equation}
V_0^\pi(s^{\high})
= \EE[\pi]{\sum_{h = 1}^H r^{\low}_{k,w,h} \mid s^{\high}_{k, w} = s^{\high}}.
\end{equation}
Here, thanks again to Approximation \ref{appr:hier_structure}, we can omit the subscripts $k$ and $w$ in $V_h$.
Equivalently, the value functions are the unique solution to the Bellman equations
\begin{equation}
\begin{split}
V_h^\pi(s^{\high}, a^{\high}, s^{\low})
&= \sum_{s^{\prime} \in \mathcal{S}} P_{h}\left(s^{\prime}; s^{\high}, a^{\high}, s^{\low}, \pi_h(s^{\high}, a^{\high}, s^{\low}) \right) V_{h+1}^\pi\left(s^{\high}, a^{\high}, s^{\prime}\right)\\
&\qquad + R_{h}(s^{\high}, a^{\high}, s^{\low}, \pi_h(s^{\high}, a^{\high}, s^{\low}) ), \qquad
\textnormal{ for } h \geq 1,\\
V_0^\pi(s^{\high})
&= \sum_{s^{\prime} \in \mathcal{S}} P_0\left(s^{\prime}; s^{\high}, \pi_0(s^{\high}) \right) V_1^\pi\left(s^{\high}, \pi_0(s^{\high}), s^{\prime}\right),
\end{split}
\end{equation}
where we set $V_{H+1}^\pi=0$. 
The optimal value function is $V_h^*(s)=\max _{\pi \in \Pi} V_h^\pi(s)$.

The goal is to maximize $\mathbb{E}_{\mathrm{Alg}}\left[\sum_{k=1}^K \sum_{w=1}^W V_0^{\pi^{k,w}}\left(s_{k,w,0}\right)\right]$, where ``Alg" denotes an RL algorithm. 
The cumulative expected regret incurred by the algorithm over $K$ episodes and $W$ time blocks is
\begin{equation}
\operatorname{Regret}(K, W, \mathrm{Alg})=\mathbb{E}_{\mathrm{Alg}}\left[\sum_{k=1}^K \sum_{w=1}^W \p{V_0^*\left(s_{k,w,0}\right)-V_0^{\pi^{k,w}}\left(s_{k,w,0}\right)}\right].
\end{equation}
where the expectation is taken over both the randomness in the algorithm and the randomness in the rewards and state transitions. 

Finally, we define 
\begin{equation}
    N_{k,w}(s) = \sum_{\substack{\breve{k} < k \text{ or}\\ (\breve{k}=k \text{ and } \breve{w}<w)}} \mathbbm{1}\cb{s^{\high}_{\breve{k},\breve{w}} = s}
\end{equation}
to be the total number of instances when the high-level state equals $s$ prior to time block $w$ and episode $k$. We use $N_{k,w}(s)$ to set hyperparameters in the following theorem. 

Now that we have established the necessary notation and definitions, we can proceed to state our main regret bound.
\begin{theo}[Regret Bound]
\label{theo:regret_bound}
    Assume the reward is bounded in $[0,1]$, the action spaces and state spaces are finite discrete, and the feature mappings are one-hot encodings. Furthermore, assume that Approximations \ref{appr:episodic_mdp} and \ref{appr:hier_structure} hold true. 
    Define 
    \begin{equation}
        S = \abs{\mathcal{S}^{\high}}
        \big(\abs{\mathcal{A}^{\high}}+1\big) \big(\abs{\mathcal{S}^{\low}}+1\big),
        \quad
        A = \abs{\mathcal{A}^{\high} \cup \mathcal{A}^{\low}}.
    \end{equation}
    Then, Algorithm \ref{alg:rlsvi_greedy_more_episodes} with the following choice of parameters:
    \begin{equation}
        \sigma^2_{k,w} = \sigma^2_{\TS,k,w} = \frac{1}{2} H^3 S \log \p{2 H S A N_{k,w}(s^{\high}_{k,w})},
    \end{equation}
    and 
    \begin{equation}
                \lambda_{k,w} = \lambda_{\TS,k,w}
                = 1/\sigma^2_{k,w}.
    \end{equation}
has cumulative expected regret bounded by
\begin{equation}
    \operatorname{Regret}\left(K, W, \operatorname{Algorithm} \ref{alg:rlsvi_greedy_more_episodes}\right) \leq \tilde{O}\left(H^3 S^{3/2} A^{1/2} \abs{\mathcal{S}^{\high}}^{1/2} \sqrt{KW}\right).
\end{equation}
The notation $\tilde{O}$ ignores poly-logarithmic factors in $H$, $S$, $A$, $\abs{\mathcal{S}^{\high}}$, $K$ and $W$. 
\end{theo}

Theorem \ref{theo:regret_bound} establishes a $\sqrt{KW}$ regret bound of the dyadic RL algorithm. Notably, this is a regret bound sublinear in $KW$, which implies that the learned policy will converge to the optimal policy when $K$ goes to infinity.

Note that the regret bound above depends on the size of both the state and action spaces. Regarding its dependency on the size of the state space, this bound may not be the tightest possible. Our results are based on \citep{russo2019worst}, who noted that the application of optimistic algorithms could yield tighter regret guarantees than RLSVI \citep{azar2017minimax, dann2017unifying}. Additionally, there are papers that provide improved regret bounds \citep{agrawal2021improved} or regret bounds in settings with function approximation \citep{zanette2020frequentist} for modified versions of RLSVI. Our choice to utilize the regret bound from \citep{russo2019worst} is not because it provides the tightest regret bound, but because it allows us to integrate RLSVI into our algorithm without any modifications, thereby ensuring interpretability.

Finally, we wish to comment on the role of the hierarchical structure in the regret bound. For now, let's set aside the dependency on the size of the state and action spaces and concentrate solely on the dependency of the regret bound on $H$, $K$, and $W$. The regret of our dyadic RL algorithm is bounded by $\tilde{O}(H^3\sqrt{KW})$. Without the knowledge of the hierarchical structure, that is, if we do not adopt Approximation 2, one might apply the RLSVI directly to the entire trial with a horizon of $HW$. In such scenarios, using the same proof technique, the regret bound becomes $\tilde{O}((HW)^3\sqrt{K})$. It's evident that $(HW)^3\sqrt{K}$ is significantly larger than $H^3\sqrt{KW}$, especially when $W$ is large. Thus, from a theoretical standpoint, leveraging the hierarchical structure aids in reducing regret.

\section{Simulations}
\label{section:simulations}

In this section, we conduct simulations to demonstrate the performance of the proposed dyadic RL algorithm in comparison to several baseline algorithms, full RL (Algorithm \ref{alg:full_RL}), stationary RLSVI (Algorithm \ref{alg:stan_RLSVI}) and a bandit (Algorithm \ref{alg:bandit_baseline}); see Section \ref{subsection:baselines} for further details. We use a toy game-like environment to illustrate the relative performance of these RL algorithms across settings with different reward structures and degree of delayed effects. In Section \ref{section:application}, we will examine the performance of our proposed algorithm within a more realistic setting, wherein we construct a test bed based on prior mobile health data. Code for reproducing the experiments is available at
\url{https://github.com/StatisticalReinforcementLearningLab/DyadicRLCode}.

\subsection{Environment}
\begin{figure}
    \centering
    \includegraphics[width = 0.8\textwidth]{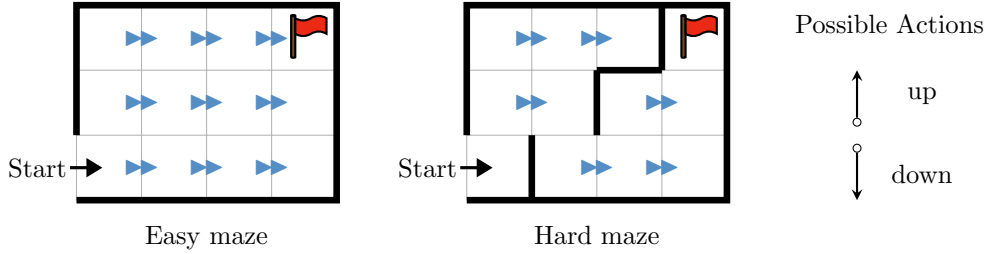}
    \caption{Two different types of mazes: there is a drift towards right, and the action set is $\cb{\operatorname{up}, \operatorname{down}}$. }
    \label{fig:mazes}
\end{figure}

An agent plays a game that is episodic in nature, comprising $K=100$ episodes. Each episode is divided into $W=15$ rounds (time blocks). At the beginning of a round, the agent observes the high-level state, the ``weather" at that particular round and must decide whether to traverse one of the two possible mazes (refer to Figure \ref{fig:mazes} for the structure of the two mazes). Following the selection of a maze, the agent can take up to a maximum of $H=7$ steps (time periods), potentially accruing rewards throughout this journey. Once the 7 steps are completed, the round concludes. The agent cannot earn any further rewards from this round, and a new round commences. This type of gaming structure similar to Boda Borg (\texttt{https://www.bodaborg.com/}), a chain of entertainment facilities. Here, visitors arrive, select specific ``quests", and attempt to complete them within a fixed timeframe.

More specifically, as depicted in Figure~\ref{fig:mazes}, the easy maze presents no obstacles, while the hard maze contains certain barriers. These obstacles complicate navigation towards the goal in the hard maze. The agent receives greater rewards when navigating the hard maze compared to the easy maze. In each round, the agent always begins in the bottom left corner of the maze. The agent can choose to move upward or downward, though there is a slight chance that it may not end up moving in the intended direction. There is a consistent rightward drift: at the end of each step, there is a high probability that the agent is shifted one unit to the right, provided no obstacle is present there. The agent's movements are more random when the weather is bad.

In reference to the terminology we have used in the paper: 
the weather condition corresponds to the high-level state, while the choice between the easy and hard maze is the high-level action. The low-level state refers to the agent's coordinates within the maze, and the low-level action is the decision to move up or down. Different reward structures are employed within the maze: generally, as the agent approaches the goal (indicated by the red flag), it earns larger rewards.

\subsubsection{Details of the construction of the environment}

Denote $p$ as the probability of the agent moving as intended. The probability that the agent moves vertically in the intended direction is $p$, while the probability of them not moving vertically or moving in the opposite direction is $(1-p)/2$ each. If any obstacles prevent the agent from moving upward/downward, the respective probability gets transferred to the ``non-moving" category. The probability of the agent moving rightward is also $p$, provided there is no obstacle on the right. If an obstacle is present, they remain in the same location along the horizontal axis. The goal state is absorbing and once reached no more rewards can be obtained. If the weather is good, we assign $p = 0.9$; conversely, in bad weather, we set $p = 0.6$.

\subsubsection{Variants of the toy environments}

\begin{figure}
    \centering
    \includegraphics[width = 0.8\textwidth]{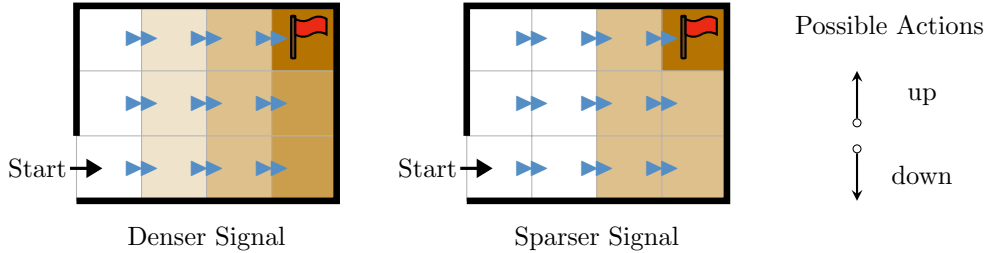}
    \caption{Two different reward structures.}
    \label{fig:mazes_reward}
\end{figure}

\paragraph{Reward structures}
We consider two distinct reward structures.
Each location within the maze is assigned a score, as depicted in Figure \ref{fig:mazes_reward}. In the left panel of Figure \ref{fig:mazes_reward}, the scores range from 0 to 4, corresponding to color gradations from white to dark brown. In contrast, the right panel of Figure \ref{fig:mazes_reward} assigns scores of 0 to 2, also correlated with color gradations from white to dark brown.
The reward at step $h$ is defined as $r$ times the difference in scores, i.e., score at step $h+1$ minus score at step $h$. It is worth noting that the agent exclusively progress towards positions of higher score (due to rightward drift); thus, their score throughout a given round never decreases. 
The reward multiplier $r$ varies based on the agent's choice of maze. If the agent chooses the easy maze, $r$ is set to 1. Conversely, if the agent opts for the hard maze, $r$ is increased to 1.2.

\paragraph{Delayed effect of high-level actions}
\label{subsection:simulation_delayed_setting}
We incorporate a delayed effect of the high-level actions. The underlying intuition is as follows: if the agent consistently chooses the hard maze, the agent becomes ``tired", causing its movements to become more random in the upcoming mazes.

To be more precise, in the $w$-th round (time block), we define:
\begin{equation}
    \text{Tiredness}_w = \sum_{l = 1}^{w - 1} 0.5^{w-l} a^{\high}_{l}.
\end{equation}
This represents an exponentially weighted sum of past instances of opting for the hard maze. Consequently, we adjust the probability $p$, i.e., the probability of the agent moving as intended, based on the value of $\text{Tiredness}_w$. Specifically, we take:
\begin{equation}
    p = \begin{cases}
        1 - 0.3\times \text{Tiredness}_w \times\tau^{\operatorname{delayed}} \qquad & \text{ if weather is good},\\
        0.7 - 0.3\times \text{Tiredness}_w \times\tau^{\operatorname{delayed}} \qquad & \text{ if weather is bad}. 
    \end{cases} 
\end{equation}
We consider different values of $\tau^{\operatorname{delayed}}$: 1/3 (weak delayed effect), 2/3 (medium delayed effect), 1 (strong delayed effect). 

In our experiments, we consider five distinct toy environments, each differing in terms of reward structure and the degree of delayed effects involved (Table \ref{table:toy_environments}).

\begin{table}
\centering
\begin{tabular}{|
>{\columncolor[HTML]{EFEFEF}}c |c|c|c|c|}
\hline
  & \cellcolor[HTML]{EFEFEF}Reward & \cellcolor[HTML]{EFEFEF}Delayed Effect & \cellcolor[HTML]{EFEFEF}$p$: good weather & \cellcolor[HTML]{EFEFEF}$p$: bad weather \\ \hline
Toy Environment 1 & Dense & No & 0.9 & 0.6\\ \hline
Toy Environment 2 & Sparse & No & 0.9 & 0.6\\ \hline
Toy Environment 3 & Sparse & Weak & $1 - 0.1\times \text{Tiredness}_w$ & $0.7 - 0.1 \times \text{Tiredness}_w$ \\ \hline
Toy Environment 4 & Sparse & Medium & $1 - 0.2\times \text{Tiredness}_w$ & $0.7 - 0.2 \times\text{Tiredness}_w$ \\ \hline
Toy Environment 5 & Sparse & Strong & $1 - 0.3\times \text{Tiredness}_w$ & $0.7 - 0.3 \times\text{Tiredness}_w$ \\ \hline
\end{tabular}
\caption{Summary of the toy environments considered.}
\label{table:toy_environments}
\end{table}

\subsubsection{Some properties of the environment}
\label{subsection:environment_prop}
Note that within a round (time block) the environment is not a bandit environment, which is particularly evident with the reward structure depicted in the right panel of Figure \ref{fig:mazes_reward}. In this scenario, the agent would always receive a zero reward for their initial step, yet the optimal policy is to attempt to move up.

Secondly, the optimal policy should depend on the time period number in the hard maze. Consider, for example, the position $(3,3)$, which is a dead-end: once the agent enters $(3,3)$, it becomes trapped, thereby eliminating any potential for earning further rewards. Consequently, in the early stages of a round, if the agent were to act optimally, it would ideally avoid entering $(3,3)$. However, as the round draws to a close, moving from $(2,3)$ to $(3,3)$ could present an opportunity to secure some additional, quick rewards, making the dead-end location potentially favorable towards the end.

Both of these features align with the real-world context encountered in mobile health studies. For example, in ADAPTS HCT, we expect that the action on one day will impact the state on the next day and thus the within-week environment is not a bandit environment. 
We will further elaborate on these points in Section~\ref{section:application}.

\subsection{Implementation details and baseline algorithms}
\label{subsection:baselines}
The dyadic RL algorithm (Algorithm~\ref{alg:rlsvi_greedy_more_episodes}) makes uses of the hierarchical structure of the problem. Specifically, it recognizes the similarities across different rounds (time blocks), which enables learning within an episode. Without this understanding of the hierarchical structure, an RL algorithm could only learn between episodes, but not within an episode across different rounds. Bearing this in mind, we consider two baseline episodic RL algorithms that do not employ the hierarchical approach---a full RL algorithm and a stationary RLSVI algorithm. Another natural baseline algorithm to consider is a bandit algorithm, which essentially regards each time period as a separate episode. To summarize, we have the following three baseline algorithms for comparison. 

\begin{enumerate}
    \item Full RL (Algorithm \ref{alg:full_RL}). The full RL algorithm implements RLSVI with horizon $T=7 \times 15$ and learns between the $K=100$ episodes. On the first day of each time block, the algorithm observes both the high-level and low-level states and decides on both high-level and low-level actions accordingly.
    \item Stationary RLSVI (Algorithm \ref{alg:stan_RLSVI}). The stationary RLSVI algorithm, much like the full RL algorithm, only learns between episodes and implements the stationary RLSVI algorithm as introduced in \citep{osband2016generalization}. It modifies the action space in a manner similar to the full RL algorithm. 
    \item Bandit (Algorithm \ref{alg:bandit_baseline}). We implement the Thompson Sampling algorithm. Again, it modifies the action space in a manner similar to the full RL algorithm. 
\end{enumerate}
See Appendix~\ref{appendix:addi_alg} for details of these algorithms. 

In all the aforementioned algorithms, one-hot encoding is utilized for feature mapping. 
To warm start all the algorithms, we use a Bernoulli distribution with a parameter of 0.5 to randomize both the high-level and low-level actions for the first episode. The parameters $\lambda, \lambda_{\TS}, \sigma, \sigma_{\TS}$ are all chosen to be 1. 

We anticipate slower convergence for the Full RL algorithm due to the large set of parameters it maintains when implementing RLSVI. Both the bandit algorithm and the stationary RLSVI algorithm learn policies that do not depend on the time period number $h$. As such, neither is expected to converge to the optimal policy.

\subsection{Results}

\subsubsection{Without delayed effect at the level of time blocks}
\begin{figure}
     \centering
     \begin{subfigure}[b]{0.4\textwidth}
         \centering
         \includegraphics[width=\textwidth,trim={25 0 25 0},clip]{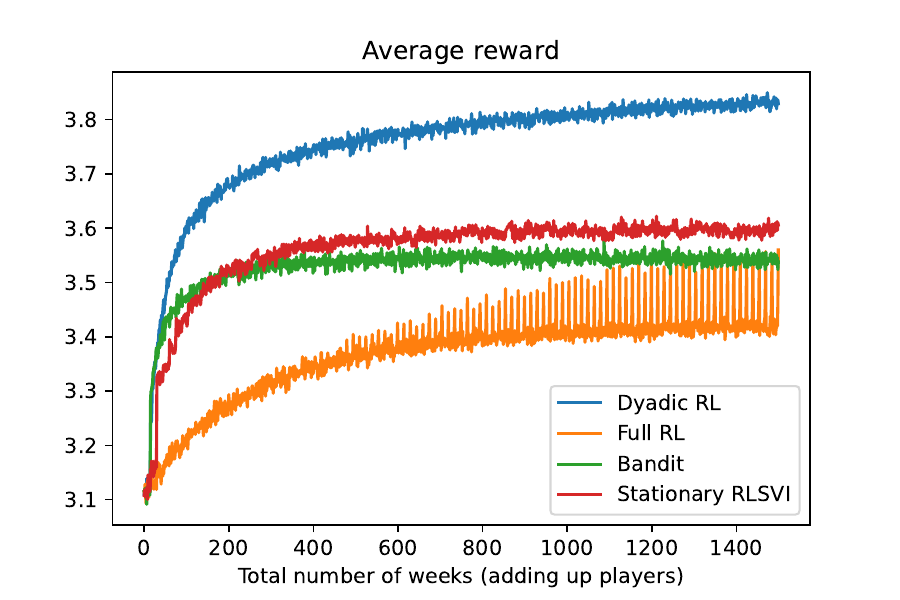}
         \caption{Expected sum of reward in a time block}
         \label{fig:reward_more_bandit}
     \end{subfigure}
     \begin{subfigure}[b]{0.4\textwidth}
         \centering
         \includegraphics[width=\textwidth,trim={25 0 25 0},clip]{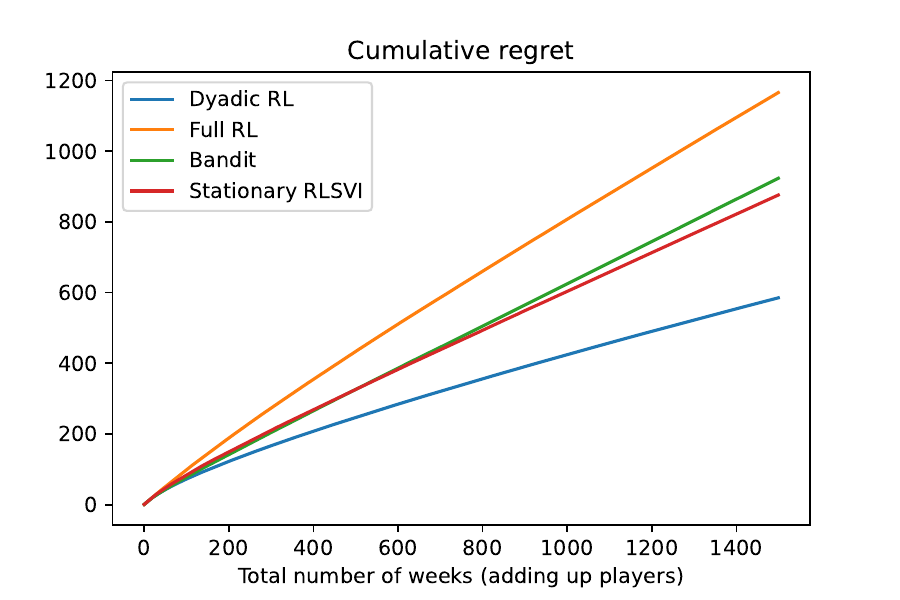}
         \caption{Cumulative regret}
         \label{fig:regret_more_bandit}
     \end{subfigure}
        \caption{Toy environment 1 in Table \ref{table:toy_environments}: A simulation setting where the reward signal is denser. There is no delayed effect at the level of time blocks. The results are an average of 10,000 independent experimental repetitions.}
        \label{fig:bandit_setting}
\end{figure}

\begin{figure}
     \centering
     \begin{subfigure}[b]{0.4\textwidth}
         \centering
         \includegraphics[width=\textwidth,trim={25 0 25 0},clip]{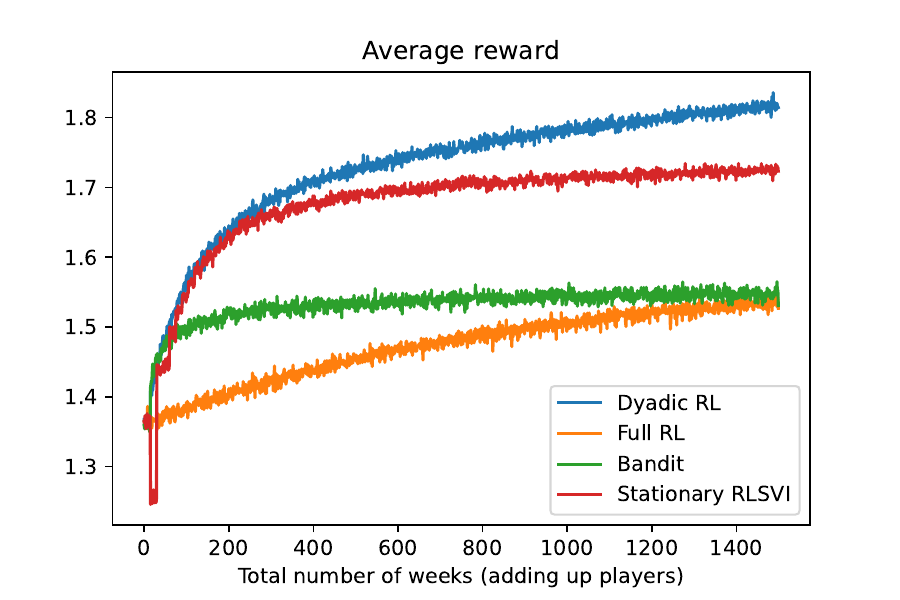}
         \caption{Expected sum of reward in a time block}
         \label{fig:reward_none_bandit}
     \end{subfigure}
     \begin{subfigure}[b]{0.4\textwidth}
         \centering
         \includegraphics[width=\textwidth,trim={25 0 25 0},clip]{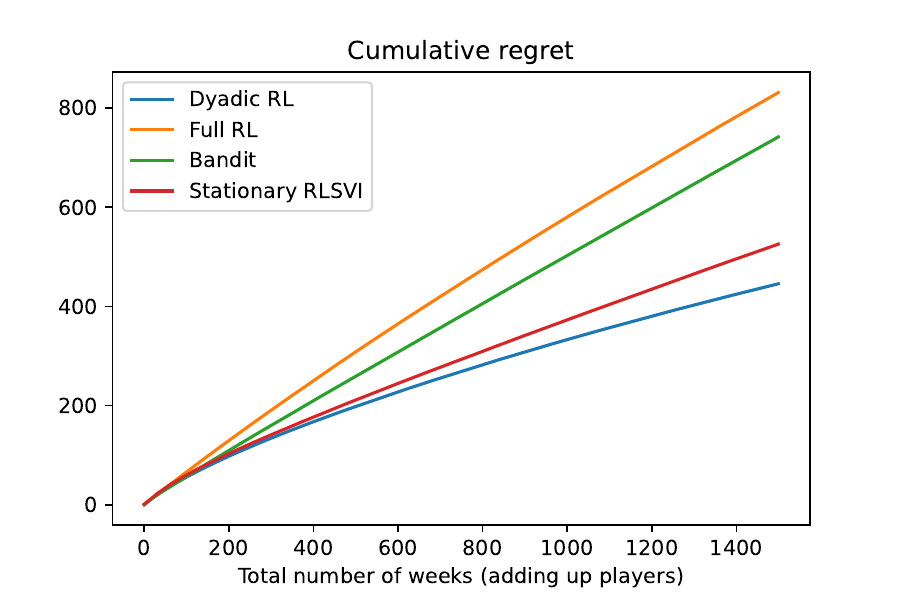}
         \caption{Cumulative Regret}
         \label{fig:regret_none_bandit}
     \end{subfigure}
        \caption{Toy environment 2 in Table \ref{table:toy_environments}: A simulation setting where the reward signal is sparser. There is no delayed effect at the level of time blocks. The results are an average of 10,000 independent experimental repetitions.}
        \label{fig:nonbandit_setting}
\end{figure}

We start with examining settings without delayed effects at the level of time blocks (Toy environments 1 and 2 in Table \ref{table:toy_environments}). The proposed dyadic RL algorithm is executed alongside the three baseline algorithms referenced in Section~\ref{subsection:baselines}. In Figure~\ref{fig:bandit_setting}, we employ the ``denser signal" reward structure depicted in the left panel of Figure~\ref{fig:mazes_reward}. The ``sparser signal" reward structure is considered in Figure~\ref{fig:nonbandit_setting}. The expected sum of rewards in a time block is illustrated in Figures~\ref{fig:reward_more_bandit} and \ref{fig:reward_none_bandit}, with the ``expectation" being the average of 10,000 independent experiment repetitions. The cumulative regret over time blocks is plotted in Figures~\ref{fig:regret_more_bandit} and \ref{fig:regret_none_bandit}.

The presented plots illustrate that our proposed dyadic RL algorithm outperforms all other algorithms evaluated, achieving higher rewards and lower regret. It is observable from the reward plot that the bandit algorithm displays a rapid initial learning phase, only to unfortunately converge to a wrong policy later. A similar phenomenon can be noticed in the stationary RLSVI algorithm's learning curve---it initially exhibits a fast learning speed, yet ultimately converges to an incorrect policy. This behavior can be attributed to the fact that the optimal policy should depend on the time period number in a time block. Detailed discussions on the environmental properties can be found in Section \ref{subsection:environment_prop}. In contrast, the full RL algorithm demonstrates a significantly slower learning pace. This is attributable to its approach of considering an entire episode of a game as a single episode in the episodic algorithm, leading to the maintenance of an excessive number of parameters. This analysis underscores the benefits of implementing a hierarchical algorithm in such a setting.

Interestingly, a ``comb" pattern emerges within the reward curve of the full RL algorithm. Upon closer inspection, it becomes clear that the full RL algorithm performs more effectively in the final time block of each episode. This performance anomaly is partially attributable to our construction of $X$ and $y$ in RLSVI. Since earlier instances of $y$ within an episode are constructed based on subsequent instances, the latter ones exhibit substantially less noise when compared to their earlier counterparts. 

\subsubsection{With delayed effect at the level of time blocks}

\begin{figure}
     \centering
     \begin{subfigure}[b]{0.32\textwidth}
         \centering
         \includegraphics[width=\textwidth,trim={25 0 20 0},clip]{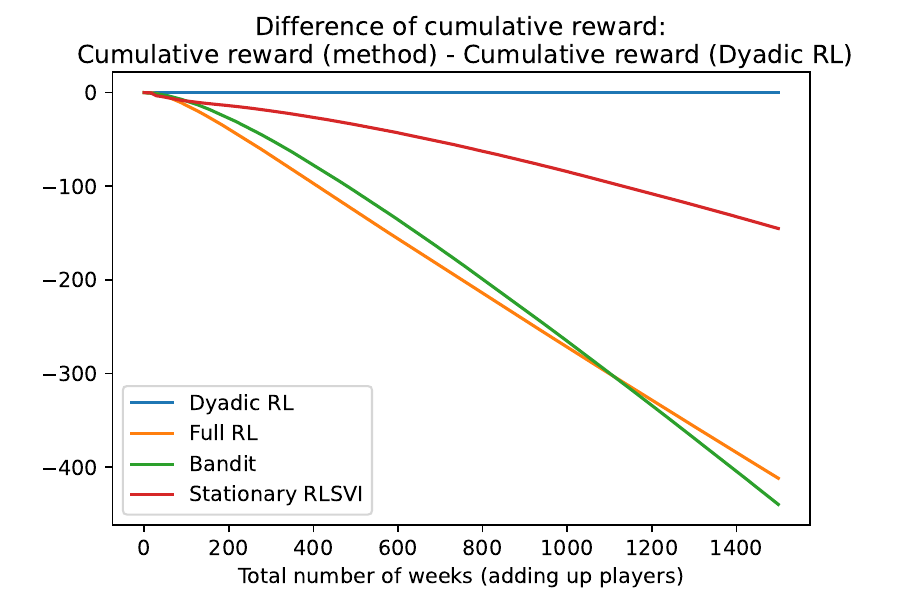}
         \caption{Weak delayed effect}
         \label{fig:regret_weak}
     \end{subfigure}
     \hfill
     \begin{subfigure}[b]{0.32\textwidth}
         \centering
         \includegraphics[width=\textwidth,trim={25 0 20 0},clip]{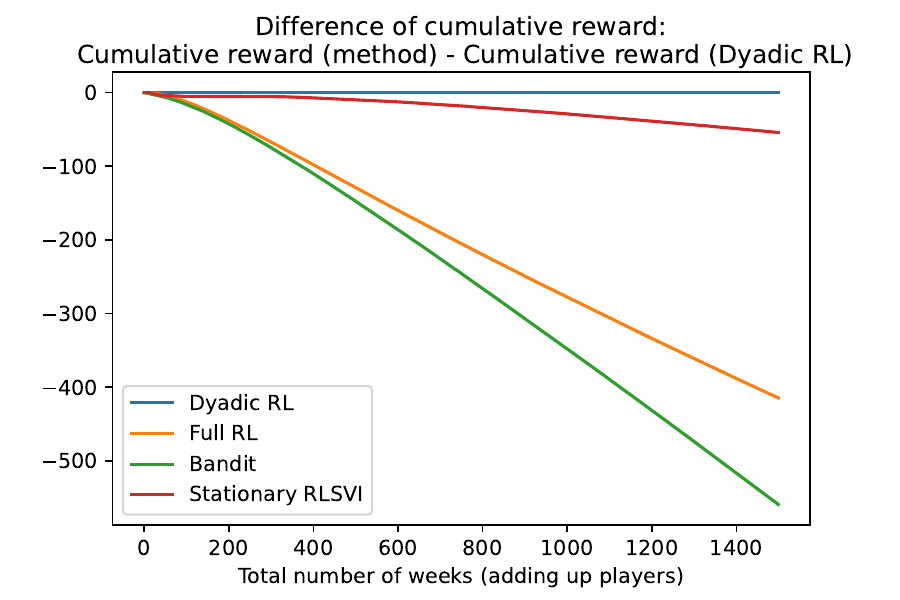}
         \caption{Medium delayed effect}
         \label{fig:regret_medium}
     \end{subfigure}
     \hfill
     \begin{subfigure}[b]{0.32\textwidth}
         \centering
         \includegraphics[width=\textwidth,trim={25 0 20 0},clip]{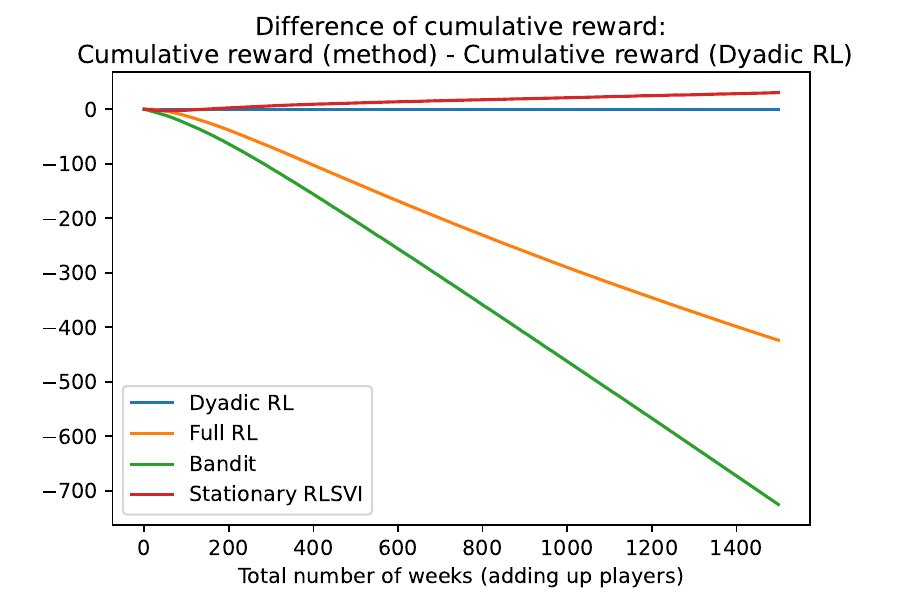}
         \caption{Strong delayed effect}
         \label{fig:regret_strong}
     \end{subfigure}
        \caption{Toy environments 3-5 in Table \ref{table:toy_environments}: Differences of the cumulative rewards between baselines and the dyadic RL algorithm. The environments have different levels of delayed effects. The results are an average of 10,000 independent experimental repetitions.}
        \label{fig:delayed_effect_regret}
\end{figure}

We next turn our attention to settings with delayed effects at the level of time blocks (Toy environments 3-5 in Table \ref{table:toy_environments}). As before, our focus remains on the proposed dyadic RL algorithm and the three baseline algorithms mentioned in Section~\ref{subsection:baselines}. The ``sparser signal" reward structure, demonstrated in the right panel of Figure~\ref{fig:mazes_reward}, is utilized in our investigation. In Figure~\ref{fig:delayed_effect_regret}, we plot the differences between the cumulative regret of the baseline algorithms and the cumulative regret of the dyadic RL algorithm. We consider three different strengths of delayed effects, as detailed in Section \ref{subsection:simulation_delayed_setting}. 

As can be seen from the plots, the dyadic RL algorithm still outperforms the other three baseline algorithms in a majority of settings. This illustrates the robustness of the proposed algorithm against weak and medium delayed effects. As these delayed effects intensify, the bandit algorithm's performance deteriorates, while the full RL algorithm and the stationary RLSVI algorithm show signs of improvement. Ultimately, with the presence of strong delayed effects, the stationary RLSVI algorithm slightly outperforms our method. 

To delve deeper into this phenomenon, let's consider the problem through the lens of the variance-bias tradeoff. Essentially, both the bandit algorithm and the stationary RLSVI algorithm pull data from all time periods for learning, while the dyadic RL algorithm keep parameters separate for each time period within a time block. The full RL algorithm only pulls data from episodes, keeping parameters separate for each time period and each time block. Therefore, 
\begin{equation}
\label{eqn:variance}
\Var{\text{Bandit}} \approx 
\Var{\text{Stationary RLSVI}} < 
\Var{\text{Dyadic RL}} < 
\Var{\text{Full RL}}.
\end{equation}
When it comes to bias, the relationships are inverted:
\begin{equation}
\label{eqn:bias}
\Bias{\text{Bandit}} >
\Bias{\text{Stationary RLSVI}} > 
\Bias{\text{Dyadic RL}} > 
\Bias{\text{Full RL}}.
\end{equation}
In scenarios with strong delayed effects, the bandit, stationary RLSVI, and dyadic RL algorithms all exhibit high bias. This scenario obscures the bias advantage of the dyadic RL algorithm over the stationary RLSVI algorithm. Given the lower variance of the stationary RLSVI algorithm, it tends to perform better. The reason the full RL algorithm does not excel in our settings is that its variance is excessively high.

\section{A Simulation Test Bed Built on the Roadmap 2.0 Study}
\label{section:application}

In this section, we discuss the design of a simulation test bed, drawing on prior data from a study of the Roadmap 2.0 app. Using this test bed, we evaluate the performance of our proposed dyadic RL algorithm.

The primary objective of this test bed is to replicate the noise level and structure that is typically encountered in practical applications. Real-world environments often feature a highly complex noise structure, encompassing unobserved variables, time-correlated residuals, and potential model misspecifications. These factors contribute significantly to the overall complexity of the problem.

Code for reproducing the experiments is available at
\url{https://github.com/StatisticalReinforcementLearningLab/DyadicRLCode}. The data can be accessed at \url{https://github.com/StatisticalReinforcementLearningLab/Roadmap2.0Testbed}. 

\subsection{Design of the simulation test bed}
We leverage data from a study focusing on Roadmap 2.0, which is a care partner-facing mobile health application \citep{rozwadowski2020promoting}. Roadmap 2.0 involves 171 dyads, each consisting of a patient undergoing HCT (target person) and a care partner. 
Each member of the dyad had the Roadmap mobile app on their smartphone and wore a Fitbit wrist tracker. 
The Fitbit wrist tracker recorded physical activity, heart rate, and sleep patterns. Furthermore, each participant was asked to self-report their mood via the Roadmap app every evening. The data comprises minute-by-minute heart rate records, sleep duration at various timestamps, step count at various timestamps, and daily mood score. 

The above Roadmap 2.0 data appears ideal for constructing a simulation test bed for use in developing the RL algorithm for ADAPTS HCT in that Roadmap 2.0 concerns patients with cancer undergoing HCT and their caregivers. However, from the viewpoint of evaluating dyadic RL algorithms with the longer term goal of deploying dyadic RL in ADAPTS HCT, this data is impoverished \citep{trella2022designing}. Roadmap 2.0 does not include daily intervention actions (i.e., whether to send a motivational prompt to the patient) nor weekly intervention actions (i.e., whether to encourage the dyad to play the joint game). Thus, to build the simulation testbed, we use Roadmap 2.0 data to construct the reward and state to state transition functions under no intervention (action coded as $0$). Then to complete the test bed, we add in effects of the actions (see below).

\subsubsection{State and reward construction under no actions}

The steps in constructing the reward and states from Roadmap 2.0 involve preprocessing the Roadmap 2.0 data, specifically aggregating the data to form weekly and daily states and standardizing the states. Next, we learn individual specific base models for the reward and state to state transition under no action (i.e., the environment). 

First, we aggregate the data at the daily level to form the low-level states. In particular, on day $t$, for the $i$-th participant (target person or care partner), we get
\[
\begin{split}
\heart_{i,t} &= \textnormal{Average heart rate from 8am on day $t-1$ to 8am on day $t$},\\
\sleep_{i,t} &= \textnormal{Total sleep time from 8pm on day $t-1$ to 8am on day $t$},\\
\step_{i,t} &= \textnormal{Total step count from 8am on day $t-1$ to 8am on day $t$}.
\end{split}
\]
We set the low-level states to be daily heart rate, daily sleep time, and daily step count. 
Further, we work with the step count at a square root scale, because the distribution of the step count is right skewed: $\sqrtstep_{i,t} = \sqrt{\step_{i,t}}$. 
We utilize only the initial 15 weeks of data for each participant, translating to the first 105 days. We must also contend with missing data at the daily aggregated level. To mitigate this, we exclude dyads that exhibit a missing rate exceeding 75\% for any variable within these initial 15 weeks. This filtering process leaves us with 49 dyads.
Following the filtering process, we perform missing value imputation: we implement the mice algorithm \citep{van2011mice} for each individual participant, effectively filling in the missing data. We take the average mood of the preceding week to be the high-level state; for participant $i$ in week $w$, we form
\[
\weeklymood_{i,w} = \textnormal{Average score of mood in week $w-1$}. 
\]

The simulated study will span 14 weeks. It is designed as though a weekly mood measurement was taken even before the commencement of day 1. Thus, we will utilize data from Roadmap 2.0 starting from day 8, supplemented by $\weeklymood$ from the end of week 1 in Roadmap 2.0. In the testbed, day 8 from Roadmap 2.0 will be redefined as day 1, and $\weeklymood_{i,2}$ will be recast as $\weeklymood_{i,1}$.

We standardize all variables. In particular, to standardize a variable, we use the time specific average and the time specific standard deviation across the same type of individual (target person or care partner).
From now on, when we refer to the variables $\heart_{i,t}, \sleep_{i,t}, \sqrtstep_{i,t}, \weeklymood_{i,w}$, we mean their standardized values.

\paragraph{Base model of the environment}
In this environment, a time block corresponds to a week, and a time period corresponds to a day. Each week consists of 7 days (time periods). An episode corresponds to a dyad. The high-level actions are the weekly interventions and the low-level actions are the daily interventions. 
For a dyad pair of (target person, care partner) $(i,j)$, we take the low-level state at day $h$ in week $w$ to be
\[s^{\low}_{i,w,h} = \p{\heart_{i,7(w-1)+h}, \sleep_{i,7(w-1)+h}, \sqrtstep_{i,7(w-1)+h}}.\]
We take the high-level state at the beginning of week $w$ to be the weekly mood of both the target person and the care partner:
\[s^{\high}_{i,w} = \p{\weeklymood_{i,w}, \weeklymood_{j,w}}.\]
The daily reward is the square root of the daily step count of the target person:
\[
r^{\low}_{i,w,h} = \sqrtstep_{i,7(w-1)+h + 1}. 
\]

We model $\heart_{i,t}, \sleep_{i,t}, \sqrtstep_{i,t}$ with generalized estimating equations \citep{hardin2012generalized}. In particular, for each participant $i$ (with $c(i)$ being the other person in the dyad) and each daily variable, we fit a separate linear model with the AR(1) working correlation using the GEE approach \citep{hojsgaard2006r}:
\[
\begin{split}
\heart_{i,t+1} &\sim \heart_{i,t} + \sleep_{i,t} + \sqrtstep_{i,t} + \weeklymood_{i,w(t)}+ \weeklymood_{c(i),w(t)},\\
\sleep_{i,t+1} &\sim \heart_{i,t} + \sleep_{i,t} + \sqrtstep_{i,t} + \weeklymood_{i,w(t)}+ \weeklymood_{c(i),w(t)},\\
\sqrtstep_{i,t+1} &\sim \heart_{i,t} + \sleep_{i,t} + \sqrtstep_{i,t} + \weeklymood_{i,w(t)}+ \weeklymood_{c(i),w(t)}.\\
\end{split}
\]
Here, $w(t)$ is the number of week day $t$ is in. We extract coefficients and residuals from each one of the fitted models. Let $\beta$ denote the fitted coefficients and $\varepsilon$ denote the residuals:
\begin{equation}
\label{eqn:daily_states}
\begin{split}
\heart_{i,t+1} &= \p{1, \heart_{i,t} , \sleep_{i,t} , \sqrtstep_{i,t} , \weeklymood_{i,w(t)}, \weeklymood_{c(i),w(t)}}^\top \beta_{\heart,i} + \varepsilon_{\heart,i,t+1},\\
\sleep_{i,t+1} &= \p{1, \heart_{i,t} , \sleep_{i,t} , \sqrtstep_{i,t} , \weeklymood_{i,w(t)}, \weeklymood_{c(i),w(t)}}^\top \beta_{\sleep,i} + \varepsilon_{\sleep,i,t+1},\\
\sqrtstep_{i,t+1} &= \p{1, \heart_{i,t},\sleep_{i,t},\sqrtstep_{i,t}, \weeklymood_{i,w(t)},\weeklymood_{c(i),w(t)}}^\top \beta_{\sqrtstep,i} + \varepsilon_{\sqrtstep,i,t+1}.
\end{split}
\end{equation}

We model the high-level states similarly. For each participant $i$, we fit a separate linear model with the AR(1) working correlation using the GEE approach \citep{hojsgaard2006r} and extract coefficients and residuals: 
\[
\weeklymood_{i,w+1} \sim \weeklymood_{i,w} + \weeklymood_{c(i),w}.
\]
\begin{equation}
\label{eqn:weekly_states}
\weeklymood_{i,w+1} = \p{1, \weeklymood_{i,w},\weeklymood_{c(i),w}}^\top \theta_{\weeklymood,i} + \eta_{\weeklymood,i,w+1}.
\end{equation}

\subsubsection{Adding in effects of the intervention actions}
Recall that Roadmap 2.0 does not contain either daily or weekly intervention actions.
In this test bed we add intervention effects only to $\sqrtstep$ with heart rate and the sleep time not directly impacted by the daily and weekly intervention actions. Nonetheless, the effects of the intervention actions will propagate to heart rate and sleep time through the impact of the actions on the step count (See \eqref{eqn:daily_states} for more details). To add the effects of the daily and weekly interventions to $\sqrtstep$, we set 
\[
\begin{split}
\sqrtstep_{i,t+1} &= \p{1, \heart_{i,t},\sleep_{i,t},\sqrtstep_{i,t}, \weeklymood_{i,w(t)},\weeklymood_{c(i),w(t)}}^\top \beta_{\sqrtstep,i}\\
& \qquad \qquad \qquad \qquad \qquad \qquad \qquad \qquad \qquad \qquad+ \tau^{\high}_{i,t}a^{\high}_{i,w(t)} + \tau^{\low}_{i,t}a^{\low}_{i,t} + \varepsilon_{\sqrtstep,i,t+1}.
\end{split}
\]
Here $\tau^{\high}_{i,t}$ tracks the effect of the weekly interventions and $\tau^{\low}_{i,t}$ tracks the effect of the daily interventions. 

We truncate the variables to ensure they remain within a reasonable range, even under the influence of an ``extreme" treatment effect. Specifically, we set the following constraints: $\heart_{i,t} \in [55,120], \sleep_{i,t} \in [0, 43200], \sqrtstep_{i,t} \in [0,200], \weeklymood_{i,w(t)}\in [0,10]$.

We start with daily interventions, $\tau^{\low}_{i,t}$. We construct simulated participants with heterogeneous treatment effect related to the value of $\beta_{\sqrtstep,i}$. We also add an intervention burden effect. For day $t$, assume that it corresponds to day $h$ in week $w$. Let $\burden_{i,t} = (1-\gamma)\sum_{s = 0}^{h-1} a_{i,t-s} \gamma^{s}$ be an exponentially weighted sum of the past interventions received in the week. The use of a burden effect is to mimic the real life problem that if a participant receives too many interventions (e.g., prompts on their smart device), they may disregard subsequent prompts or even disengage from the app. 
Here we set the shrinkage factor $\gamma = 1-1/7$. We choose this shrinkage factor because it corresponds to an effective decision-making time horizon of 7 days. 
Define
\begin{equation}
\label{eqn:low_effect}
\tau^{\low}_{i,t}
= \p{\tau_{0,i} - \mathbbm{1}\cb{\burden_{i,t} \geq b_1} \tau_{1,i}}
\mathbbm{1}\cb{\burden_{i,s} \leq b_2 \textnormal{ for all }s\leq t\textnormal{ in the same week}}.
\end{equation}
Here, we call $b_1$ the burden threshold and $b_2$ the disengagement threshold. In \eqref{eqn:low_effect}, $\tau_{0,i}$ is the treatment effect without any burden, and $\tau_{1,i}$ the shrinkage in intervention effect when the burden is high. The term $\mathbbm{1}\cb{\burden_{i,s} \leq b_2 \textnormal{ for all }s\leq t\textnormal{ in the same week}}$ captures the behaviors of fully disengaging from the app for the rest of the week if the burden is too high. 
More specifically, we set 
\[\tau_{0,i} = \frac{1}{5}\beta_{\sqrtstep,\sqrtstep,i}, \qquad \tau_{1,i} = \frac{1}{10}\beta_{\sqrtstep,\sqrtstep,i}.\]
Here $\beta_{\sqrtstep,\sqrtstep,i}$ is the coefficient of $\sqrtstep_{i,t}$ in predicting $\sqrtstep_{i,t+1}$ in the above fitted GEE model \eqref{eqn:daily_states}. We choose this reduction in effect size because behavioral sciences suggest that the effect of any intervention is typically much smaller than that of the previous day's behavior. We vary the values of $b_1$ and $b_2$ by setting them to be 
\begin{equation}
\label{eqn:burden_thres}
b(k) = (1-\gamma) \sum_{i = 1}^k \gamma^{k-i}. 
\end{equation}
for different values of $k$ (recall $\gamma = 1-1/7$). Then, a threshold of $b(k)$ corresponds to the level of burden after receiving $k$ consecutive messages over the most recent $k$ days of the week. We provide some values of burden and their comparison with $b(k)$ in Table~\ref{table:burden_value}. 

\begin{table}
\centering
\begin{tabular}{|c|c|c|c|c|c|c|c|}
\hline
\multicolumn{6}{|c|}{Interventions} & Burden & Comparison with $b(k)$ \\ \hline
\textcolor{lightgray}{\ding{55}} &\textcolor{lightgray}{\ding{55}} &\ding{51} &\ding{51} &\ding{51} &\ding{51}
& 0.4602       & $= b(4)$       \\ \hline
\textcolor{lightgray}{\ding{55}} &\ding{51} &\ding{51} &\textcolor{lightgray}{\ding{55}} &\ding{51} &\ding{51}
& 0.4324      & $\in (b(3), b(4))$       \\ \hline
\ding{51} &\ding{51} &\textcolor{lightgray}{\ding{55}} &\textcolor{lightgray}{\ding{55}} &\ding{51} &\ding{51}
& 0.4085      &  $\in (b(3), b(4))$      \\ \hline
\textcolor{lightgray}{\ding{55}} &\textcolor{lightgray}{\ding{55}} &\textcolor{lightgray}{\ding{55}} &\ding{51} &\ding{51} &\ding{51}
& 0.3702      &  $ = b(3)$     \\ \hline
\ding{51} &\ding{51} &\ding{51} & \textcolor{lightgray}{\ding{55}}&\ding{51} &\textcolor{lightgray}{\ding{55}}
& 0.3556      & $\in (b(2), b(3))$      \\ \hline
\textcolor{lightgray}{\ding{55}} &\textcolor{lightgray}{\ding{55}} &\ding{51} & \ding{51}&\ding{51} &\textcolor{lightgray}{\ding{55}}
& 0.3174      & $\in (b(2), b(3))$      \\ \hline
\textcolor{lightgray}{\ding{55}} &\textcolor{lightgray}{\ding{55}} &\textcolor{lightgray}{\ding{55}} & \textcolor{lightgray}{\ding{55}}&\ding{51} &\ding{51}
& 0.2653      & $= b(2)$      \\ \hline
\ding{51} &\ding{51} &\textcolor{lightgray}{\ding{55}} & \ding{51}&\textcolor{lightgray}{\ding{55}} &\textcolor{lightgray}{\ding{55}}
& 0.2482      & $\in (b(1), b(2))$      \\ \hline
\textcolor{lightgray}{\ding{55}} &\textcolor{lightgray}{\ding{55}} &\ding{51} & \textcolor{lightgray}{\ding{55}}&\textcolor{lightgray}{\ding{55}} &\ding{51}
& 0.2328      & $\in (b(1), b(2))$      \\ \hline
\textcolor{lightgray}{\ding{55}} &\textcolor{lightgray}{\ding{55}} &\textcolor{lightgray}{\ding{55}} & \textcolor{lightgray}{\ding{55}}&\textcolor{lightgray}{\ding{55}} &\ding{51}
& 0.1429      & $= b(1)$      \\ \hline
\end{tabular}
\caption{Some values of burden on day 6 of a week. The discount factor $\gamma = 1-1/7$. }
\label{table:burden_value}
\end{table}

Finally, we set 
\begin{equation}
\label{eqn:high_level_on_state}
\tau^{\high}_{i,t} = \frac{1}{5}\tau_{0,i} = \frac{1}{25} \beta_{\sqrtstep,\sqrtstep,i}.
\end{equation}
When comparing $\tau^{\high}_{i,t}$ with $\tau^{\low}_{i,t}$, we see a reduction in effect size. We opt for this reduction because behavioral sciences indicate that the impact of a high-level action is generally less pronounced than that of a low-level action. In addition, the high-level action is expected to influence the state throughout the week; therefore, with the choice of 1/5 of $\tau_{0,i}$, the total effect of the high-level action on the states in a week is comparable to that of the low-level action. 

\paragraph{Variation: violation of the weekly bandit assumption}
We consider a variant of the simulation test bed in which the high-level actions (i.e., the weekly actions) affect not only subsequent daily step counts but also weekly mood (thus violating the assumption used by the high-level contextual bandit algorithm in the dyadic RL algorithm).
In this variant, the weekly intervention $a_{i,w}^{\high}$ affects both members of the dyad's weekly mood ($\weeklymood_{i,w+1}$ and $\weeklymood_{c(i),w+1}$). Recall that $\weeklymood_{i,w+1}$ corresponds to the average mood during week $w$. This is plausible considering the intervention at the start of the week (e.g., playing a game together) may be expected to positively influence the overall mood throughout the entire week, benefiting both the target person and the care partner.

We write down this effect on weekly mood as follows. 
\[
\begin{split}
\weeklymood_{i,w+1} &= \p{1, \weeklymood_{i,w},\weeklymood_{c(i),w}}^\top \theta_{\weeklymood,i}\\
& \qquad \qquad \qquad \qquad \qquad \qquad + \tau^{\operatorname{mood},\operatorname{high}}_i a_{i,w}^{\high}+ \eta_{\weeklymood,i,w+1},\\
\weeklymood_{c(i),w+1} &= \p{1, \weeklymood_{i,w}, \weeklymood_{c(i),w}}^\top \theta_{\weeklymood,c(i)}\\
& \qquad \qquad \qquad \qquad \qquad \qquad + \tau^{\operatorname{mood},\operatorname{high}}_i a_{i,w}^{\high} + \eta_{\weeklymood,c(i),w+1}.
\end{split}
\]
For a target person $i$, let $\theta_{\weeklymood,\weeklymood},i$ be the coefficient of $\weeklymood_{i,w}$ in predicting $\weeklymood_{i,w+1}$ in the previously fitted GEE model. 
We consider the following settings:
\begin{equation}
\label{eqn:delayed_effect}
\begin{split}
\tau^{\operatorname{mood},\operatorname{high}}_i &= 0 \qquad \qquad \qquad \qquad \qquad \qquad \qquad \quad \,\, \text{(No effect)},\\
\tau^{\operatorname{mood},\operatorname{high}}_i &= \frac{7}{50}\theta_{\weeklymood,\weeklymood,i} \qquad \qquad \text{ (Weak effect)},\\
\tau^{\operatorname{mood},\operatorname{high}}_i &= \frac{7}{25}\theta_{\weeklymood,\weeklymood,i}\qquad \qquad \text{ (Strong effect)},
\\
\tau^{\operatorname{mood},\operatorname{high}}_i &= \frac{14}{25}\theta_{\weeklymood,\weeklymood,i}\qquad \qquad \text{ (Extreme effect)}.
\end{split}
\end{equation}
Recall that we set $\tau_{i,t}^{\high}$ to be 1/25 (times the effect of the previous day's behavior), as outlined in \eqref{eqn:high_level_on_state}. The value of 7/25 then represents the total effect of the high-level action on the low-level states over the week. Given our expectation that the effect of the high-level action on the next week’s high-level state will be, at most, similar to its total effect on the current week’s low-level state, we define the value of 7/25 (times the effect of the previous week's behavior) as our strong effect. Furthermore, we investigate scenarios where the effect is double that of the strong effect, which we refer to as the extreme effect, as detailed in \eqref{eqn:delayed_effect}.

\subsection{Performance of algorithms}

The previous section results in 49 dyad models (environments corresponding to the dyad specific models for 49 dyads in Roadmap 2.0). To assess the performance of the algorithms (the proposed algorithm in this paper and the baselines) we simulate trials each of 100 dyads by sampling with replacement from the 49 dyads models. In the simulation, the dyads arrive in a sequential order and each of the dyad will be in the study for a total of 14 weeks, i.e, 98 days. 
For each sampled dyad, we generate the states and rewards according to the models described in the prior section. 

We apply the proposed dyadic RL algorithm to the simulation test bed, comparing it with the baseline algorithms detailed in Section \ref{subsection:baselines}: (1) Full RL, (2) Bandit, and (3) Stationary RLSVI. For implementation, the parameters $\lambda, \lambda_{\TS}, \sigma, \sigma_{\TS}$ are all set to 1. The feature mapping is taken to be the identity mapping; that is, we use linear models for the $Q$-functions when considering the lower-level agent, and linear models for the reward when considering the upper-level agent.

\begin{figure}
     \centering
     \begin{subfigure}[b]{0.31\textwidth}
         \centering
         \includegraphics[trim={60 0 50 0},clip, width=\textwidth]{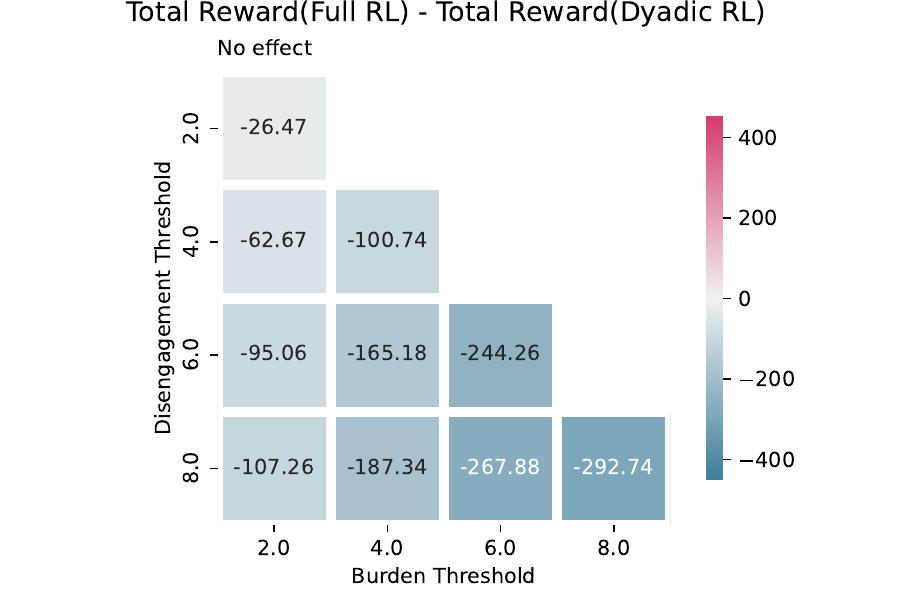}
         \caption{Full RL}
         \label{fig:real_data_no_full}
     \end{subfigure}
     \hfill
     \begin{subfigure}[b]{0.31\textwidth}
         \centering
         \includegraphics[trim={60 0 50 0},clip,width=\textwidth]{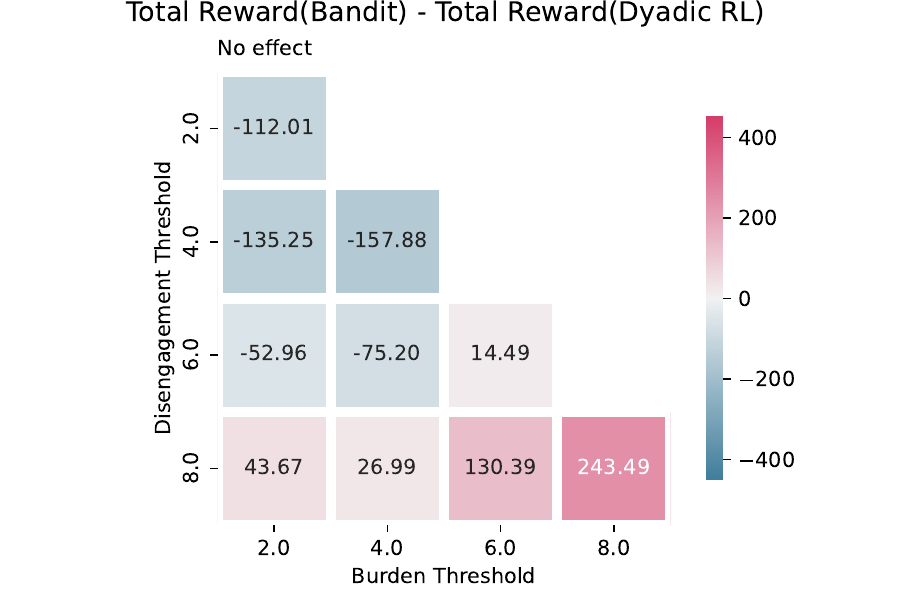}
         \caption{Bandit}
         \label{fig:real_data_no_bandit}
     \end{subfigure}
     \hfill
     \begin{subfigure}[b]{0.36\textwidth}
         \centering
         \includegraphics[trim={60 0 0 0},clip,width=\textwidth]{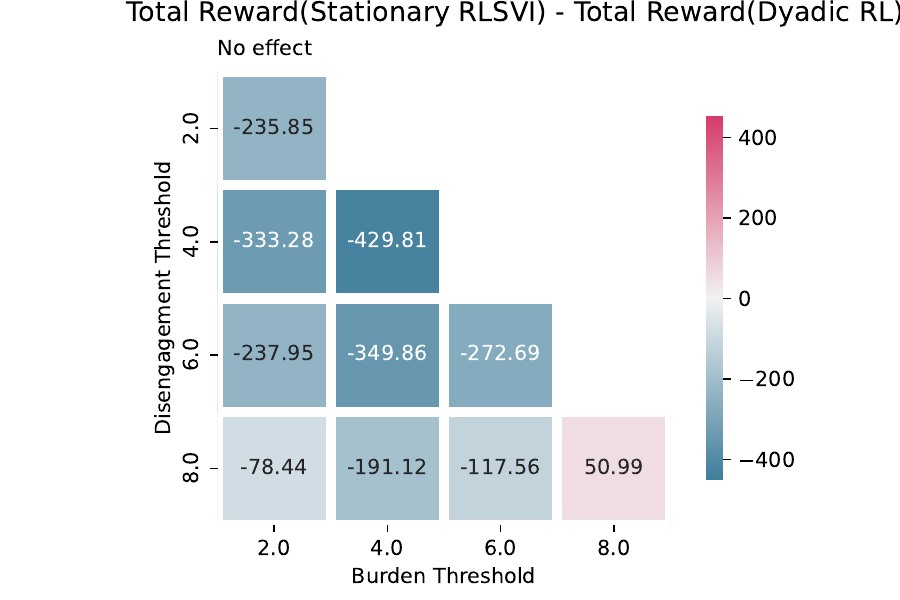}
         \caption{Stationary RLSVI}
         \label{fig:real_data_no_stan}
     \end{subfigure}
        \caption{No effect of the high-level action on subsequent time blocks' high-level state: Differences of the averaged total rewards between baselines and the dyadic RL algorithm. The $x$-axis displays the value of $k$ used to establish the burden threshold $b_1 = b(k)$; the $y$-axis represents the value of $k$ employed to set the disengagement threshold $b_2 = b(k)$. A threshold of $b(k)$ corresponds to the level of burden after receiving $k$ consecutive messages over the most recent $k$ days of the week.}
        \label{fig:real_data_no}
\end{figure}

\begin{figure}
     \centering
     \begin{subfigure}[b]{0.31\textwidth}
         \centering
         \includegraphics[trim={60 0 50 0},clip, width=\textwidth]{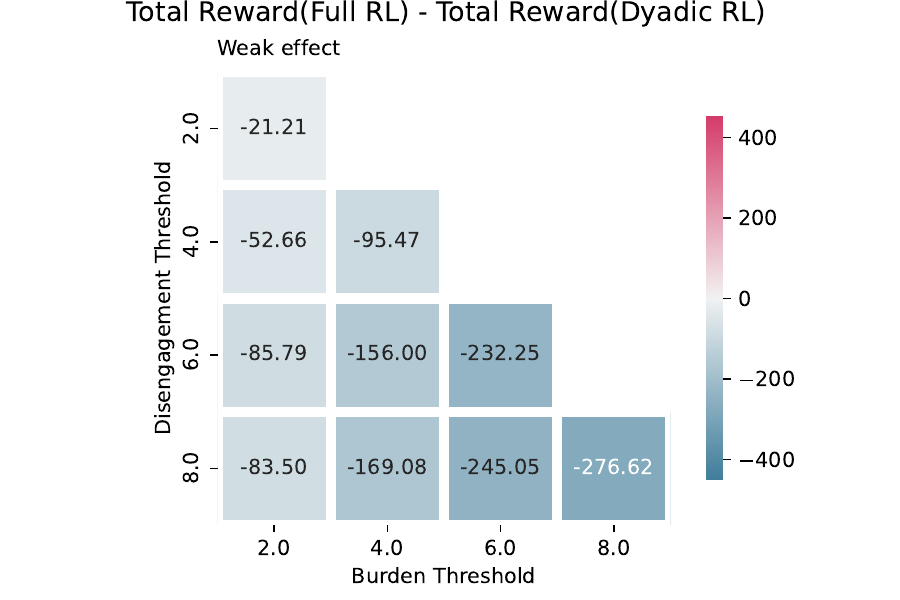}
         \caption{Full RL}
         \label{fig:real_data_weak_full}
     \end{subfigure}
     \hfill
     \begin{subfigure}[b]{0.31\textwidth}
         \centering
         \includegraphics[trim={60 0 50 0},clip,width=\textwidth]{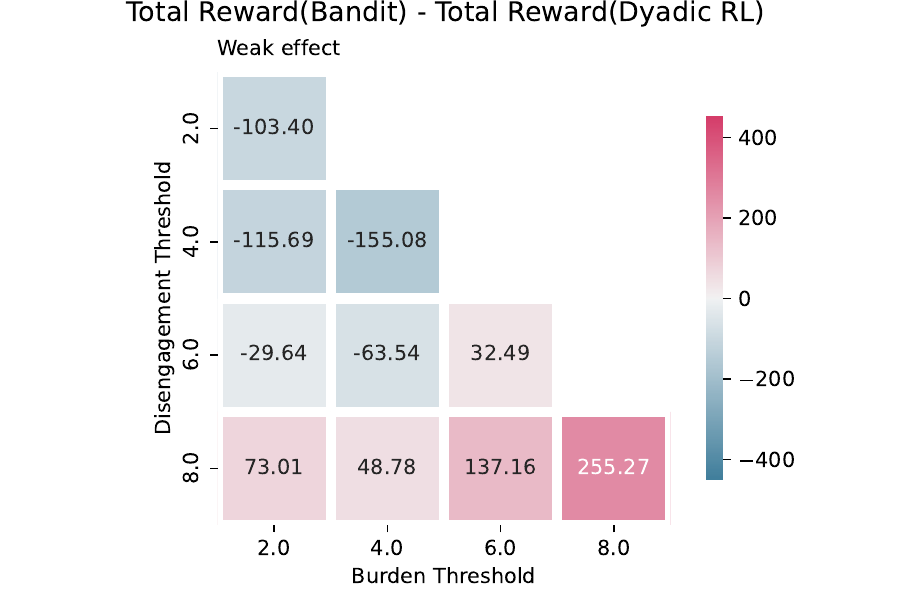}
         \caption{Bandit}
         \label{fig:real_data_weak_bandit}
     \end{subfigure}
     \hfill
     \begin{subfigure}[b]{0.36\textwidth}
         \centering
         \includegraphics[trim={60 0 0 0},clip,width=\textwidth]{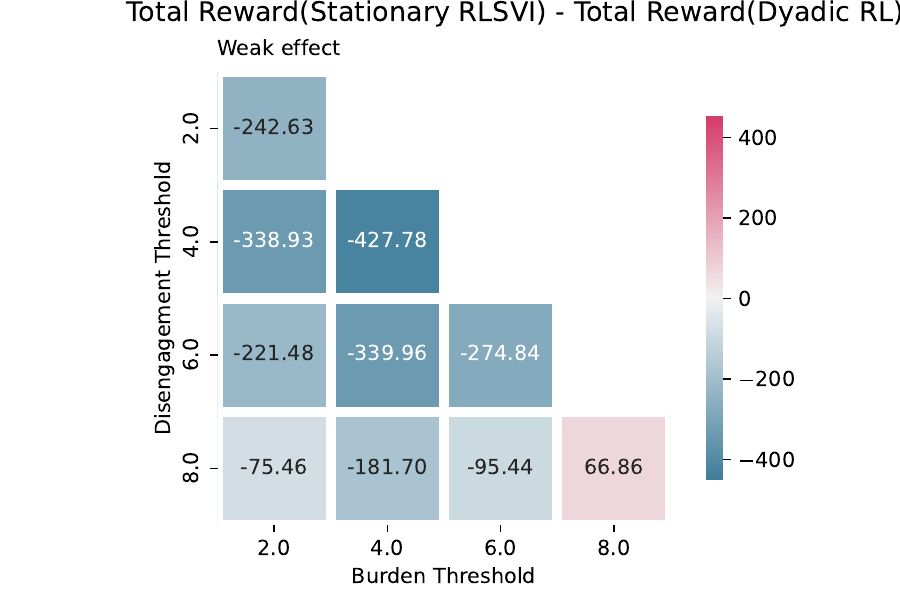}
         \caption{Stationary RLSVI}
         \label{fig:real_data_weak_stan}
     \end{subfigure}
        \caption{Weak effect of the high-level action on subsequent time blocks' high-level state: Differences of the averaged total rewards between baselines and the dyadic RL algorithm. The $x$-axis displays the value of $k$ used to establish the burden threshold $b_1 = b(k)$; the $y$-axis represents the value of $k$ employed to set the disengagement threshold $b_2 = b(k)$. A threshold of $b(k)$ corresponds to the level of burden after receiving $k$ consecutive messages over the most recent $k$ days of the week.}
        \label{fig:real_data_weak}
\end{figure}

\begin{figure}
     \centering
     \begin{subfigure}[b]{0.31\textwidth}
         \centering
         \includegraphics[trim={60 0 50 0},clip, width=\textwidth]{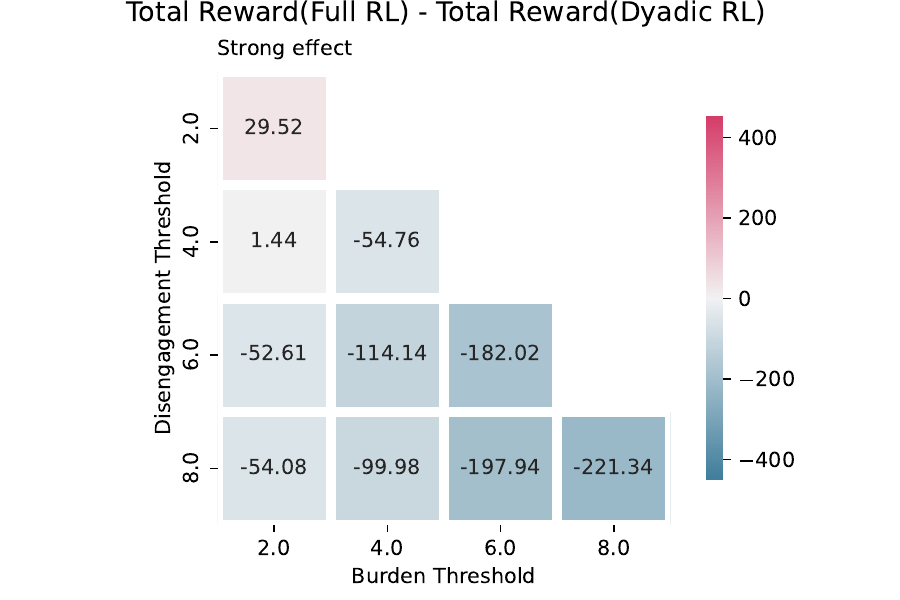}
         \caption{Full RL}
         \label{fig:real_data_strong_full}
     \end{subfigure}
     \hfill
     \begin{subfigure}[b]{0.31\textwidth}
         \centering
         \includegraphics[trim={60 0 50 0},clip,width=\textwidth]{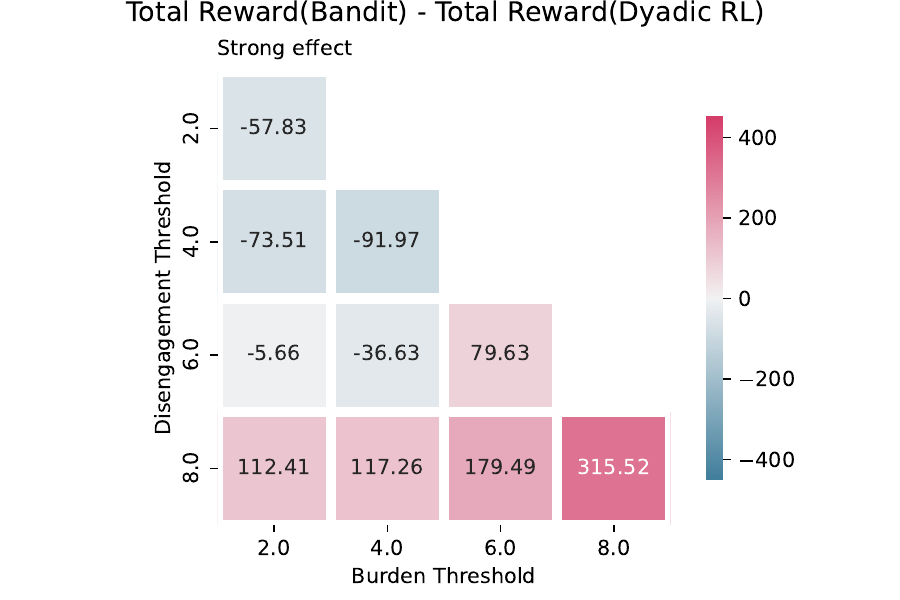}
         \caption{Bandit}
         \label{fig:real_data_strong_bandit}
     \end{subfigure}
     \hfill
     \begin{subfigure}[b]{0.36\textwidth}
         \centering
         \includegraphics[trim={60 0 0 0},clip,width=\textwidth]{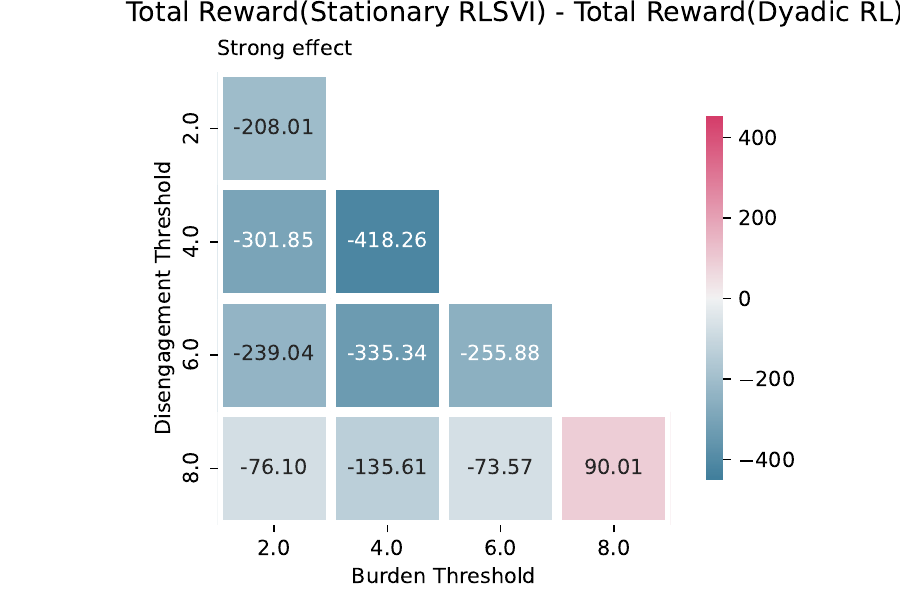}
         \caption{Stationary RLSVI}
         \label{fig:real_data_strong_stan}
     \end{subfigure}
        \caption{Strong effect of the high-level action on subsequent time blocks' high-level state: Differences of the averaged total rewards between baselines and the dyadic RL algorithm. The $x$-axis displays the value of $k$ used to establish the burden threshold $b_1 = b(k)$; the $y$-axis represents the value of $k$ employed to set the disengagement threshold $b_2 = b(k)$. A threshold of $b(k)$ corresponds to the level of burden after receiving $k$ consecutive messages over the most recent $k$ days of the week.}
        \label{fig:real_data_strong}
\end{figure}

\begin{figure}
     \centering
     \begin{subfigure}[b]{0.31\textwidth}
         \centering
         \includegraphics[trim={60 0 50 0},clip, width=\textwidth]{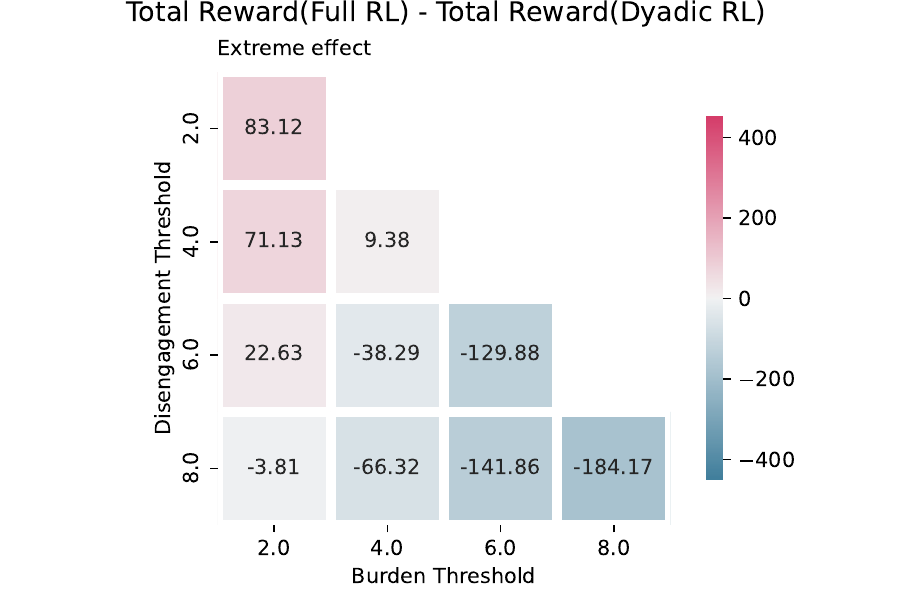}
         \caption{Full RL}
         \label{fig:real_data_extreme_full}
     \end{subfigure}
     \hfill
     \begin{subfigure}[b]{0.31\textwidth}
         \centering
         \includegraphics[trim={60 0 50 0},clip,width=\textwidth]{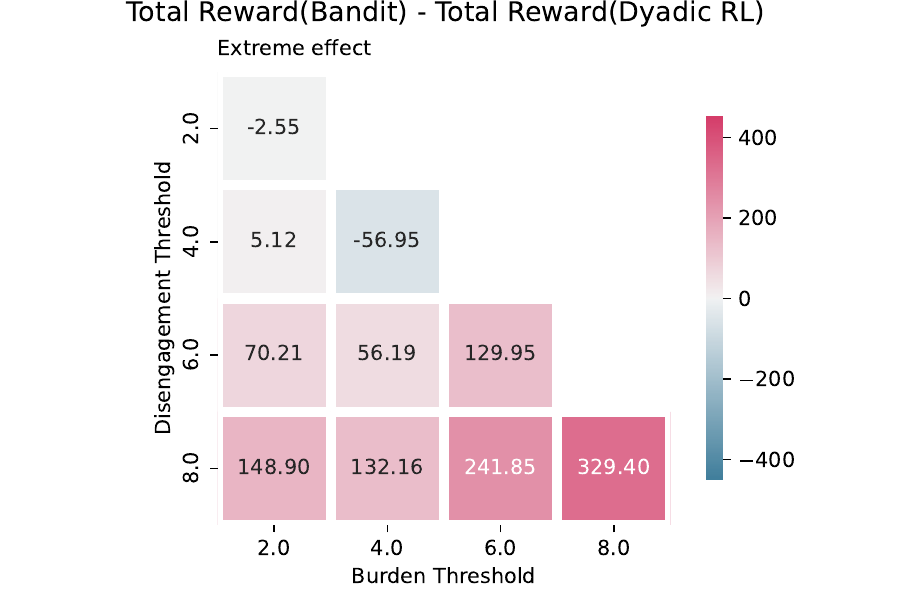}
         \caption{Bandit}
         \label{fig:real_data_extreme_bandit}
     \end{subfigure}
     \hfill
     \begin{subfigure}[b]{0.36\textwidth}
         \centering
         \includegraphics[trim={60 0 0 0},clip,width=\textwidth]{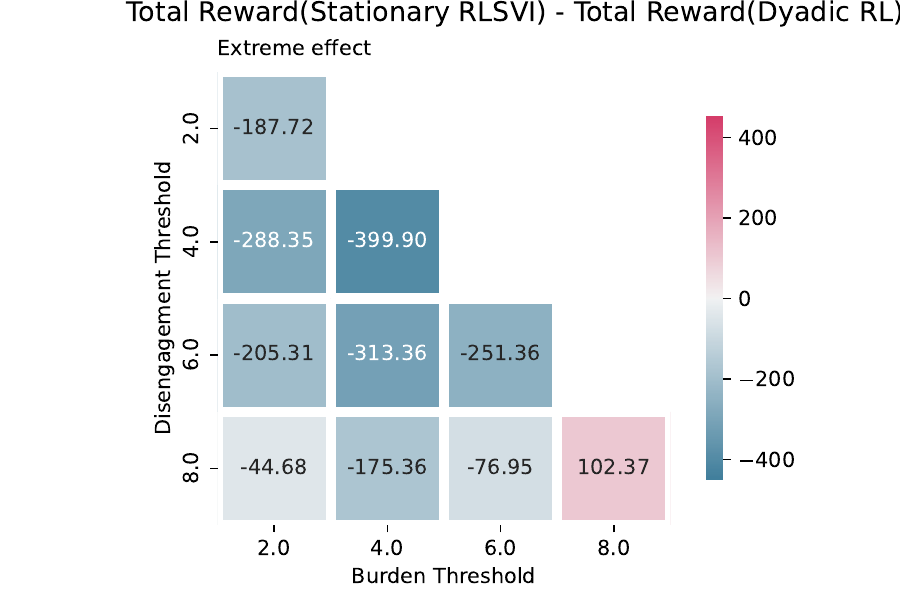}
         \caption{Stationary RLSVI}
         \label{fig:real_data_extreme_stan}
     \end{subfigure}
        \caption{Extreme effect of the high-level action on subsequent time blocks' high-level state: Differences of the averaged total rewards between baselines and the dyadic RL algorithm. The $x$-axis displays the value of $k$ used to establish the burden threshold $b_1 = b(k)$; the $y$-axis represents the value of $k$ employed to set the disengagement threshold $b_2 = b(k)$. A threshold of $b(k)$ corresponds to the level of burden after receiving $k$ consecutive messages over the most recent $k$ days of the week.}
        \label{fig:real_data_extreme}
\end{figure}

In Figures \ref{fig:real_data_no}-\ref{fig:real_data_extreme}, we illustrate the differences in averaged total rewards between the baseline algorithms and the proposed dyadic RL algorithm through heatmaps. The total reward here refers to the cumulative rewards across 14 weeks for 100 dyads. We average over 1000 simulated trials, each of 100 dyads. If the values on the plot are positive (depicted in red), then the dyadic RL algorithm has a lower averaged total reward, indicating inferior performance over the baseline algorithm.
In each subplot, the $x$-axis displays the value of $k$ used to establish the burden threshold $b_1 = b(k)$, as detailed in \eqref{eqn:burden_thres}; the $y$-axis represents the value of $k$ employed to set the disengagement threshold $b_2 = b(k)$.
Recall, $b_1$ is the threshold at which the treatment effect reduces to half, and $b_2$ is the threshold beyond which the user disengages for the week. A threshold of $b(k)$ corresponds to the level of burden after receiving $k$ consecutive messages over the most recent $k$ days of the week.

\paragraph{Effect of the high-level action on subsequent time block's high-level state: No/Weak/Strong}
When compared to the Stationary RLSVI algorithm and the full RL algorithm, the dyadic RL algorithm performs better most of the time. However, the relative performance of the full RL algorithm and the Stationary RLSVI algorithm improves when the effect of the high-level action on subsequent time blocks' high-level state is stronger. 

In comparison to the bandit algorithm, the dyadic RL algorithm outperforms when the disengagement and burden thresholds are low. This is expected since in this case there are greater delayed negative effects of the intervention. In the extreme case where there is no reduction in treatment effect due to burden or disengagement (when both thresholds are at 8), the optimal policy is to constantly send interventions. The bandit algorithm, considering only immediate rewards, converges faster to this optimal policy. Practically speaking, if domain knowledge suggests that users are highly unlikely to feel overburdened by interventions or disengage from the app, implementing a bandit algorithm could be an optimal choice.

\paragraph{Effect of the high-level action on subsequent time block's high-level state: Extreme}
When the effect of the high-level action on a subsequent time block's high-level state becomes exceedingly large, the advantages of all three baseline algorithms over the dyadic RL algorithm become more pronounced. This observation parallels our findings from the variance-bias analysis presented in \eqref{eqn:variance} and \eqref{eqn:bias}. As the ``weekly carryover" effect intensifies, all but the full RL algorithm exhibit significant bias. In such scenarios, the superiority of our algorithm in terms of bias compared to the bandit algorithm and the stationary RLSVI becomes less distinct.

\section*{Discussion}

While this paper is motivated by the planned ADAPTS HCT trial, we have not addressed all the practical challenges; our goal in this paper is to make initial steps toward the ultimate goal of developing a finalized version for use in the ADAPTS HCT trial. Here, we outline a few points where the setting considered in this paper differs from the planned ADAPTS HCT trial. 

\begin{enumerate}
\item In ADAPTS HCT, the dyads will join the trial in batches, and there will be multiple batches of dyads in the trial at any given time.
\item In ADAPTS HCT, the medication will be taken twice daily. The lower level interventions can be given at two time points each day: in the morning and in the evening.
\item In addition to the intervention targeting the adolescent and the one targeting the dyadic relationship, there is a third intervention that will target the care partner.
\item The ``week start" may not always be Sunday. Each dyad will be offered a choice of a week start day, which could include Saturday, Sunday, or possibly Monday. However, this choice would be set in advance and would remain consistent throughout the trial for each dyad.
\item The high-level action does not take the form of a “prompt at the beginning of the week to encourage game play”. At the beginning of each week, the RL algorithm will decide whether it is a non-game or game week for each dyad and communicate this via a prompt to both members of the dyad. During a non-game week, the adolescent may receive motivational messages---this could occur up to twice daily. During a game week, taking the medication provides the adolescent and parent with an opportunity to engage in the game (e.g., being able to submit a guess as part of solving a word puzzle). Motivational messages focusing on the game might be sent if the medication is not taken within a predetermined time frame. Similar to a non-game week, this could occur up to twice daily with a new puzzle offered upon solving until the end of the week. 
\end{enumerate}

\section*{Acknowledgement}
SL was supported by NIH/NIDA grant P50 DA054039 and NIH/NIDCR grant UH3 DE028723. 
SWC was supported by NIH/NHLBI grants R01 HL146354 and K24 HL156896 and NCI grant R01 CA249211.
INS was supported by NIH/NIDA grants P50 DA054039 and R01 DA039901. 
SAM was supported by NIH/NIDA grant P50 DA054039, 
NIH/NIBIB and OD grant P41 EB028242, and 
NIH/NIDCR grant UH3 DE028723.

The authors are grateful to Raaz Dwivedi, Kyra Gan, Yongyi Guo, Hsin-Yu Lai, Ivana Malenica, Anna Trella, Ziping Xu and Kelly Zhang for their constructive feedback and insightful comments.

\newpage
\bibliographystyle{plainnat}
\bibliography{ref}

\newpage
\appendix

\section{Proof of Theorem \ref{theo:regret_bound}}

We line up all blocks from various episodes in a sequence, resulting in a total of $KW$ blocks. If we concentrate on a specific high-level state $s \in \mathcal{S}^{\high}$ and extract the blocks where $s_{k,w}^{\high} = s$, we end up with a sequence of ``short" episodes within identical environments, each having a fixed horizon of $H+1$. Notably, as we employ one-hot encoding, Algorithm \ref{alg:rlsvi_greedy_more_episodes} essentially learns independently for different high-level state values. Thus, it suffices to focus on a specific $s \in \mathcal{S}^{\high}$ and examine the regret obtained on blocks where $s_{k,w}^{\high} = s$.

Next, we reformulate the problem to make the setting similar to the one in \citep{russo2019worst}. To this end, for each episode number $k$, time block number $w$ and time period number $h$, we define
\begin{alignat}{4}
\label{eqn:state_const}
    s_{k,w,0} &= \p{s_{k,w}^{\high},\NA,\NA},
    \qquad &&a_{k,w,0} = a_{k,w}^{\high},
    \qquad &&r_{k,w,0} = 0, 
    \qquad &&\text{for } h = 0,\\
    s_{k,w,h} &= \p{s_{k,w}^{\high}, a_{k,w}^{\high}, s_{k,w,h}^{\low}}, 
    \qquad &&a_{k,w,h} = a_{k,w, h}^{\low}, 
    \qquad &&r_{k,w,h} = r_{k,w, h}^{\low}, 
    \qquad &&\text{for } h = 1, \dots, H. 
\end{alignat}
Correspondingly, we define the state and action spaces to be
\begin{equation}
\mathcal{S} = \mathcal{S}^{\high} \times \p{\mathcal{A}^{\high} \cup \cb{\NA}}
 \times \p{\mathcal{S}^{\low}\cup \cb{\NA}},
 \qquad 
\mathcal{A} = \mathcal{A}^{\high} \cup \mathcal{A}^{\low}.
\end{equation}
 With this transformation, we recast the problem into a finite-horizon MDP with a horizon of $H+1$. 

Next, we delve deeper into Algorithm \ref{alg:rlsvi_greedy_more_episodes}. We notice that Algorithm \ref{alg:rlsvi_greedy_more_episodes} is fundamentally the same as applying RLSVI to the above finite-horizon MDP with a horizon of $H+1$. This is because the artificial reward given to the high-level agent, $\tilde{r}^{\high}_{\breve{k},\breve{w}} = \max_{\alpha} \tilde{\theta}_{k, w, 1} ^\top \phi_1 \left(s^{\high}_{\breve{k},\breve{w}}, a^{\high}_{\breve{k},\breve{w}}, s^{\low}_{\breve{k}, \breve{w}, 1}, \alpha \right)$, mirrors the exact form of the $Q$-function constructed in the later periods of RLSVI.

With these observations, we can directly apply the results from \citep{russo2019worst} and obtain that
\begin{equation}
    \operatorname{ConditionalRegret}(K,W,\text{Algorithm \ref{alg:rlsvi_greedy_more_episodes}}, s)
    \leq \tilde{O}\left(H^3 \abs{\mathcal{S}}^{3/2} \abs{\mathcal{A}}^{1/2} \sqrt{N_{K,W}(s)}\right),
\end{equation}
where
\begin{equation}
\begin{split}
&\operatorname{ConditionalRegret}(K,W,\mathrm{Alg}, s)
    = \\
 & \qquad \qquad \mathbb{E}_{\mathrm{Alg}}\left[\sum_{k=1}^K \sum_{w=1}^W \p{V_0^*\left(s_{k,w,0}\right)-V_0^{\pi^{k,w}}\left(s_{k,w,0}\right)} \mathbbm{1}\cb{s^{\high}_{k,w} = s} \mid s^{\high}_{0,0}, \dots, s^{\high}_{K,W}\right].
 \end{split}
\end{equation}
Here, ConditionalRegret represents the cumulative expected regret incurred on the blocks with a high-level state $s$, conditioned on all high-level states. By summing across all unique values of $s$, we obtain that
\begin{equation}
\begin{split}
 \operatorname{Regret}\left(K, W, \operatorname{Algorithm} \ref{alg:rlsvi_greedy_more_episodes}\right)
 &= \sum_{s \in \mathcal{S}^{\high}} \EE{\operatorname{ConditionalRegret}(K,W,\mathrm{Alg}, s)}\\
 & \leq \tilde{O}\left(H^3 \abs{\mathcal{S}}^{3/2} \abs{\mathcal{A}}^{1/2} \sum_{s \in \mathcal{S}^{\high}} \EE{\sqrt{N_{K,W}(s)}}\right)\\
 & \leq \tilde{O}\left(H^3 \abs{\mathcal{S}}^{3/2} \abs{\mathcal{A}}^{1/2} \abs{\mathcal{S}^{\high}}^{1/2} \EE{\sqrt{ \sum_{s \in \mathcal{S}^{\high}}N_{K,W}(s)}}\right)\\
 & = \tilde{O}\left(H^3 \abs{\mathcal{S}}^{3/2} \abs{\mathcal{A}}^{1/2} \abs{\mathcal{S}^{\high}}^{1/2} \sqrt{ KW}\right). 
\end{split}
\end{equation}
Finally, note that by definition, $S = \abs{\mathcal{S}}$, and $A = \abs{\mathcal{A}}$. 

\section{Connections to \texorpdfstring{\citet{wen2020efficiency}}{Wen et al. [2020]}}
\label{appendix:wen}

In this section, we discuss the connection of our framework to that of \citet{wen2020efficiency}. 

\citet{wen2020efficiency} focus on the problem of learning to optimize performance through repeated interactions with an unknown finite-horizon MDP $\mathcal{M}=(T, \mathcal{S}, \mathcal{A}, P, \mathcal{R}, \mu_0)$, where $\mathcal{S}$ is the state space, $\mathcal{A}$ is the action space, $P$ and $\mathcal{R}$ respectively represent the transition and reward distributions, and $\mu_0$ is the distribution of the initial state. The agent interacts with the environment across $K$ episodes. Each episode unfolds over $T$ time periods, and for each time period $t \in \cb{1,2,\dots, T}$, if the agent takes action $a_t \in \mathcal{A}$ at state $s_t \in \mathcal{S}$, then it receives a random reward drawn from distribution $\mathcal{R}(\cdot \mid s_t, a_t)$. The agent then transitions to the next state $s_{t+1} \in \mathcal{S}$ with a probability given by $P\left(s_{t+1} \mid s_t, a_t\right)$.

\citet{wen2020efficiency} employ the concept of modularity to articulate a hierarchical structure of the MDP. In particular, \citet{wen2020efficiency} define ``subMDPs" and ``equivalent subMDPs". We provide a recap of these definitions.

\begin{defi}[subMDP]
\label{defi:subMDP}
Consider a partition of the states $\mathcal{S}$ into $L$ disjoint subsets $\mathcal{H}=$ $\left\{\mathcal{S}_i\right\}_{i=1}^L$. An induced subMDP $\mathcal{M}_i=(\mathcal{S}_i \cup \mathcal{E}_i, \mathcal{A}, P_i, \mathcal{R}_i, \mathcal{E}_i)$ is:
\begin{itemize}
\item The internal state set is $\mathcal{S}_i$. The action space is $\mathcal{A}$.
\item The exit state set $\mathcal{E}_i$ is $\mathcal{E}_i=\left\{e \in \mathcal{S} \backslash \mathcal{S}_i: \exists(s, a) \in \mathcal{S}_i \times \mathcal{A}\right.$ s.t. $\left.P(e \mid s, a)>0\right\}$.
\item The state space of $\mathcal{M}_i$ is $\mathcal{S}_i \cup \mathcal{E}_i$.
\item $P_i$ and $\mathcal{R}_i$ are the restriction of $P$ and $\mathcal{R}$ to domain $\mathcal{S}_i \times \mathcal{A}$ respectively. 
\item The subMDP $\mathcal{M}_i$ terminates if it reaches a state in $\mathcal{E}_i$.
\end{itemize}
\end{defi}

\begin{defi}[Equivalent subMDPs]
\label{defi:equisubMDP}
Two subMDPs, denoted as $\mathcal{M}_i$ and $\mathcal{M}_j$, are identified as equivalent if a bijection $f: \mathcal{S}_i \cup \mathcal{E}_i \rightarrow \mathcal{S}_j \cup \mathcal{E}_j$ exists such that $f\left(\mathcal{S}_i\right)=\mathcal{S}_j, f\left(\mathcal{E}_i\right)=\mathcal{E}_j$, and, through $f$, the two subMDPs share identical transition probabilities and reward structures at their internal states. 
\end{defi}

Focusing on a partition $\mathcal{H}$ of states in $\mathcal{M}$, the set of induced subMDPs is $\cb{\mathcal{M}_i}_{i=1}^L$. Within this context, \citet{wen2020efficiency} define two key elements: $M$, representing the maximum number of states in any given subMDP, and $\mathcal{E}$, the collection of all exit states. They are defined as follows:
\begin{equation}
M=\max _i\left|\mathcal{S}_i \cup \mathcal{E}_i\right| \quad \text { and } \quad \mathcal{E}=\cup_{i=1}^L \mathcal{E}_i. 
\end{equation}
Let $l \leq L$ be the number of equivalence classes of subMDPs induced by a particular partition $\mathcal{H}$ of states in $\mathcal{M}$. 
\citet{wen2020efficiency} say that the MDP $\mathcal{M}$ exhibits hierarchical structure with respect to a partition $\mathcal{H}$, when:
\begin{equation}
M l \ll|\mathcal{S}|, \qquad \text{ and } \qquad |\mathcal{E}| \ll |\mathcal{S}|.
\end{equation}

We will now put our setting in the framework of \citet{wen2020efficiency}. 
Recall that in \citep{wen2020efficiency}, the agent interacts with a finite-horizon MDP $\mathcal{M}=(T, \mathcal{S}, \mathcal{A}, P, \mathcal{R}, \mu_0)$ across $K$ episodes. In our setting, each of the episodes is of length $T = W(H+1)$ and is composed of $\cb{s_t, a_t, r_t}_{t=1}^T$. 

We start with discussing the state. In our setting, each state is a five tuple:
\begin{equation}
\label{eqn:state_construction}
    s_t = (s_{t,1}, s_{t,2}, s_{t,3}, s_{t,4}, s_{t,5}) \in \mathcal{S},
\end{equation}
where
\begin{equation}
\label{eqn:state_space}
\begin{split}
s_{t,1} \in \mathcal{S}^w = \cb{1,2,\dots,W}&, \qquad
s_{t,2} \in \mathcal{S}^h = \cb{0,1,\dots,H}, \qquad
s_{t,3} \in \mathcal{S}^{\high}, \qquad\\
s_{t,4} \in \mathcal{A}^{\high} \cup \cb{\NA}&, \qquad
s_{t,5} \in \mathcal{S}^{\low} \cup \cb{\NA},
\end{split}
\end{equation}
and $\mathcal{S} = \mathcal{S}^w \times \mathcal{S}^h \times \mathcal{S}^{\high}  \times ( \mathcal{A}^{\high} \cup \cb{\NA}) \times (\mathcal{S}^{\low} \cup \cb{\NA})$. 
More specifically, for a state $s_t$, let $s_{t,1}$ track the time block number, $s_{t,2}$ track the time period number within the block, $s_{t,3}$ track the high-level state, $s_{t,4}$ track the high-level action, and $s_{t,5}$ track the low-level state. 
We summarize the above discussion in Table~\ref{table:notation_corres}. 

\begin{table}
\caption{Interpretation of $s_{t,1}, s_{t,2}, s_{t,3}, s_{t,4}, s_{t,5}$ in the context of ADAPTS HCT.}
\label{table:notation_corres}
\centering
\begin{tabular}{|l|l|l|l|l|l|}
\hline
State & $s_{t,1} \in \cb{1,\dots,W}$ & $s_{t,2} \in \cb{0,,\dots,H}$ & $s_{t,3} \in \mathcal{S}^{\high}$ & $s_{t,4} \in \mathcal{A}^{\high}$ & $s_{t,5} \in \mathcal{S}^{\low}$ \\ \hline
Meaning & time block $\#$ & time period $\#$ & high-level state & high-level action & low-level state\\ \hline
\end{tabular}
\end{table}

In this setting, the transition function is constrained in a variety of ways. For example, $P(s_{t+1} \mid s_t, a_t) = 0$ if any of the following holds:
\begin{enumerate}
    \item $s_{t,2} < H$ and ($s_{t+1,1} \neq s_{t,1}$, $s_{t+1,2} \neq s_{t,2} + 1$ or $s_{t+1,3} \neq s_{t,3}$);
    \item $0 < s_{t,2} < H$ and ($s_{t+1,4} \neq s_{t,4}$);
    \item $s_{t,2} = H$ and ($s_{t+1,1} \neq s_{t,1} + 1$, $s_{t+1,2} \neq 0$, $s_{t+1,4} \neq \NA$ or $s_{t+1,5} \neq \NA$). 
\end{enumerate}
    Note that the value of $s_{t,2}$ plays a pivotal role. Specifically, the states $s$ with $s_{t,2} = 0$ or $s_{t,2} = H$ resemble ``bottleneck" states as referenced in the literature, linking different densely connected regions of the state space \citep{csimcsek2004using, solway2014optimal, machado2017laplacian}. Consider $s_{t,2}$ and $s_{t,3}$ as illustrated in Figure~\ref{fig:bottleneck}: if $s_{t,2}$ is between 1 and $H-1$, then $s_{t,3}=s_{t+1,3}$. Only when the value of $s_{t,2}$ reaches $H$ may $s_{t,3}$ differ from $s_{t+1,3}$. In this scenario, we can consider the states $s$ with $s_{t,2} = 0$ or $s_{t,2} = H$ as bottleneck states (as depicted in yellow in Figure~\ref{fig:bottleneck}). Each floor or subspace of the same value of $s_{t,3}$ can be seen as a densely connected region, with the bottleneck states connecting these different regions. 

\begin{figure}
    \centering
    \includegraphics[width = 0.5\textwidth]{tikz/bottleneck.tikz}
    \caption{States $s$ with $s_{t,2} = 0$ or $s_{t,2} = H$ are considered as bottleneck states: Only when the value of $s_{t,2}$ reaches $H$ is $s_{t,3}$ allowed to change at the next transition. Each floor or subspace of the same value of $s_{t,3}$ can be seen as a densely connected region, with the bottleneck states connecting these different regions.}
    \label{fig:bottleneck}
\end{figure}

We then move on to discuss the action. In our setting, there are two types of actions: a high-level action $a^{\high} \in \mathcal{A}^{\high}$ and a low-level action $a^{\low} \in \mathcal{A}^{\low}$. Using notation from \citet{wen2020efficiency}, we define $\mathcal{A} = \mathcal{A}^{\high} \cup \mathcal{A}^{\low}$. Recall that we allow the effects of the high-level state and action to last throughout the time block, and thus when constructing states in \eqref{eqn:state_construction}, we incorporate the high-level state and action into $s_{t,3}$ and $s_{t,4}$ to make the environment Markovian. 
At the beginning of each time block, the action chosen by the RL algorithm is the high-level action. During the time block, at each time period, the action chosen by the RL algorithm is the low-level action.

A reward is incurred at each time period within each time block. We assume that at the beginning of time block, after the high-level action is chosen, the reward is always zero.

Returning to our discussion of bottleneck states in Figure \ref{fig:bottleneck}, each time block can be treated as a densely connected region, and the states $s$ with $s_{t,2}=0$ and $s_{t,2} = H$ resemble the ``bottleneck" states. In the context of ADAPTS HCT, the state $s$ with $s_{t,2}=0$ might correspond to a Sunday. Note also that the concept of ``bottleneck" states is closely connected to the concept of exit states \citep{wen2020efficiency}. Therefore, it is logical to define subMDPs based on time blocks and treat any state with $s_{t,2} = 0$ as an exit state. Specifically, we partition the state space $\mathcal{S}$ based on the time block number $w$ and obtain $\mathcal{H} = \cb{\mathcal{S}_w}_{w=1}^W$. 

Using the above notation, Approximation \ref{appr:hier_structure} translates to the following two properties: 
\begin{proper}[Exit state]
\label{proper:exit_state}
There exists a function $\nu$ such that
for any $s_t$ with $s_{t,2} = H$, the transition probability satisfies:
\begin{equation}
P(s_{t+1} \mid s_t, a_t) = \nu(s_{t+1}; s_{t,1}, s_{t,3}). 
\end{equation}
\end{proper}
In words, the above property indicates that the transition probability to any exit state does not depend on the past action, $s_{t,4}$ (high-level action) or $s_{t,5}$ (low-level state). 
Instead, it relies solely on $s_{t,3}$ (high-level state). This introduces a stronger concept of a ``bottleneck" state because this property further limits the degree of connections between subMDPs.

\begin{proper}[subMDP equivalence]
\label{proper:sub_equi}
All subMDPs are equivalent. 
\end{proper}
Property \ref{proper:sub_equi} enables our algorithm to learn across time blocks. If the time blocks were non-equivalent, then the RL algorithm can only learn between episodes.

The above two properties imply that there is a hierarchical structure in the environment. Suppose that $\mathcal{S}^{\high}$, $\mathcal{S}^{\low}$, $\mathcal{A}^{\high}$, and $\mathcal{A}^{\low}$ are finite. Then the size of each subMDP, $\mathcal{S}_w$, is upper bounded by $(H+1) \abs{\mathcal{S}^{\high}} \abs{\mathcal{A}^{\low}} \abs{\mathcal{S}^{\low}}$. Consequently, the value of $M$, which represents the maximum size of any subMDP and the set of all exit states, is upper bounded by $(H+2) \abs{\mathcal{S}^{\high}} \abs{\mathcal{A}^{\low}} \abs{\mathcal{S}^{\low}}$. Following Property \ref{proper:sub_equi}, there is only $l = 1$ equivalence class in total. As for the size of the state space, if we allow for the time-blocks to be fully non-equivalent, then $\abs{\mathcal{S}} \approx H W \abs{\mathcal{S}^{\high}} \abs{\mathcal{A}^{\low}} \abs{\mathcal{S}^{\low}}$. Finally, the size of the set of all exit states is $\abs{\mathcal{E}} = W \abs{\mathcal{S}^{\high}}$. Therefore, we have that
\begin{equation}
    M l \leq (H+2) \abs{\mathcal{S}^{\high}} \abs{\mathcal{A}^{\low}} \abs{\mathcal{S}^{\low}}
    \ll H W \abs{\mathcal{S}^{\high}} \abs{\mathcal{A}^{\low}} \abs{\mathcal{S}^{\low}} \approx \abs{\mathcal{S}} ,
\end{equation}
if the number of time blocks $W$ is large, and that
\begin{equation}
 \abs{\mathcal{E}} = W \abs{\mathcal{S}^{\high}} 
 \ll H W \abs{\mathcal{S}^{\high}} \abs{\mathcal{A}^{\low}} \abs{\mathcal{S}^{\low}} \approx \abs{\mathcal{S}},
\end{equation}
if $H \abs{\mathcal{A}^{\low}} \abs{\mathcal{S}^{\low}}$ is large. 
Thus, in line with our discussion in Section \ref{subsection:appr_envi}, a hierarchical structure exists, and it is thus desirable to take advantage of such structure to develop an RL algorithm. 

\section{Additional Algorithms}
\label{appendix:addi_alg}
In this section, we present detailed algorithm boxes for the baseline algorithms studied in Sections \ref{section:simulations} and \ref{section:application}. See Section \ref{subsection:baselines} for a high-level, intuitive discussion on the algorithms. 

\begin{algorithm}
\caption{Thompson Sampling (Generic)} 
\label{alg:thompson_sampling_generic}
\begin{algorithmic}
\Require Data $\cb{s_1, a_1, r_{1}, \dots, s_{l-1}, a_{l-1}, r_{l-1}}$;
Feature mapping $\psi$; Parameters $\lambda_{\TS}, \sigma_{\TS}>0$.
\begin{enumerate}
\item Generate regression problem $X \in \mathbb{R}^{(l-1) \times p}, y \in \mathbb{R}^{l-1}$ :
\[
X = \left[\begin{array}{c}
\psi\left(s_1, a_1\right)^\top \\
\vdots \\
\psi\left(s_{l-1}, a_{l-1}\right) ^\top
\end{array}\right], \qquad 
y_{\breve{l}} = r_{\breve{l}} \textnormal{ for } \breve{l} = 1, \dots, l-1. 
\]

\item Bayesian linear regression for the value function:
\[\begin{array}{l}
\bar{\beta}_{l} \leftarrow \frac{1}{\sigma_{\TS}^2}\left(\frac{1}{\sigma_{\TS}^2} X^{\top} X + \lambda_{\TS} I\right)^{-1} X^{\top} y, \\
V_{l} \leftarrow\left(\frac{1}{\sigma_{\TS}^2} X^{\top} X+\lambda_{\TS} I\right)^{-1}.
\end{array}\]

\item Sample $\tilde{\beta}_{l} \sim N\left(\bar{\beta}_{l}, V_{l}\right)$ from Gaussian posterior. 
\end{enumerate}
\Ensure $\tilde{\beta}_{l}$. 
\end{algorithmic}
\end{algorithm}


\begin{algorithm}
\caption{Full RL with RLSVI (Episodic)}
\label{alg:full_RL}
\begin{algorithmic}
\Require Feature mapping $\phi$; Parameters $\sigma>0, \lambda>0$. 
\State $\text{Data} = \cb{}$. 
\For{Episode $k = 1, 2, \dots, K$}
\State 
Input $H = H W$, $\text{Data}$ and $\sigma, \lambda$ into Algorithm \ref{alg:rlsvi_generic} and get output $\tilde{\theta}_{k,1,1}, \dots, \tilde{\theta}_{k,w, H}$. 
\For{Time block $w = 1, 2, \dots, W$}

\State Observe $s_{k,w}^{\high}$.

\For{Time period $h = 1, \dots, H$}
\State Observe $s_{k,w,h}^{\low}$.
\If{ $h = 1$}
\State Sample $\p{a_{k,w}^{\high}, a_{k,w, h}^{\low}}\in \operatorname{argmax}_{\alpha, \alpha'} \tilde{\theta}_{k,w, h} ^\top \phi \left(s^{\high}_{k,w}, \alpha, s^{\low}_{k,w, h}, \alpha' \right)$.
\Else
\State Sample $a_{k,w, h}^{\low} \in \operatorname{argmax}_{\alpha} \tilde{\theta}_{k,w, h} ^\top \phi \left(s^{\high}_{k,w}, a^{\high}_{k,w}, s^{\low}_{k,w, h}, \alpha \right)$.
\EndIf
\State Observe $r^{\low}_{k,w,h}$.
\State Add $(s_{k, (w-1)H + h}, a_{k, (w-1)H + h}, r^{\low}_{k, (w-1)H + h})$ to Data with
\[
\begin{split}
s_{k, (w-1)H + h} &= \p{s_{k,w}^{\high}, a_{k,w}^{\high}, s_{k,w,h}^{\low}}, \qquad r_{k, (w-1)H + h} = r^{\low}_{k,w,h},\\
 a_{k, (w-1)H + h} &= 
  \begin{cases}
     \p{a_{k,w}^{\high}, a_{k,w,h}^{\low}} &\text{ if } h = 1,\\
     a_{k,w,h}^{\low} &\text{ otherwise}.
 \end{cases}
 \end{split}
 \]
\EndFor
\EndFor
\EndFor
\end{algorithmic}
\end{algorithm}

\begin{algorithm}
\caption{Stationary Randomized Least-Squares Value Iteration (Generic)}
\label{alg:stan_rlsvi_generic}
\begin{algorithmic}
\Require 
Data $\cb{s_{\breve{l},1}, a_{\breve{l},1}, r_{\breve{l},1}, \dots, s_{\breve{l},H}, a_{\breve{l},H}, r_{\breve{l},H}}_{\breve{l} = 1\dots, l-1}$;
Feature mapping $\phi$;
Previous estimate $\tilde{\theta}_{l-1}$; Parameters $\lambda, \sigma>0, \gamma \in [0,1]$.
\begin{enumerate}
\item Generate regression problem $X \in \mathbb{R}^{(H(l-1)) \times p}, y \in \mathbb{R}^{H(l-1)}$ :
\[\begin{array}{l}
X = \left[\begin{array}{c}
\phi\left(s_{1,1}, a_{1,1} \right) ^\top \\
\phi\left(s_{1,2}, a_{1,2} \right) ^\top \\
\vdots \\
\phi\left(s_{l-1, H}, a_{l-1, H}\right) ^\top
\end{array}\right], \\
\quad \\
y_{(\breve{l}-1)H + h} = \left\{\begin{array}{ll}
r_{\breve{l}, h}+ \gamma \max _\alpha \tilde{\theta}_{l -1 }^\top \phi\left(s_{\breve{l}, h+1}, \alpha\right) & \text { if } h < H, \\
r_{\breve{l}, h} & \text { if } h = H.
\end{array}\right.
\end{array}\]

\item Bayesian linear regression for the value function:
\[\begin{array}{l}
\bar{\theta}_{l} \leftarrow \frac{1}{\sigma^2}\left(\frac{1}{\sigma^2} X^{\top} X + \lambda I\right)^{-1} X^{\top} y, \\
\Sigma_{l} \leftarrow\left(\frac{1}{\sigma^2} X^{\top} X+\lambda I\right)^{-1}.
\end{array}\]

\item Sample $\tilde{\theta}_{l} \sim N\left(\bar{\theta}_{l}, \Sigma_{l}\right)$ from posterior.
\end{enumerate}
\Ensure $\tilde{\theta}_{l}$
\end{algorithmic}
\end{algorithm}


\begin{algorithm}
\caption{Stationary RLSVI (Episodic)}
\label{alg:stan_RLSVI}
\begin{algorithmic}
\Require Feature mapping $\phi$; Parameters $\sigma>0, \lambda>0$. 
\State $\text{Data} = \cb{}$. 
\For{Episode $k = 1, 2, \dots, K$}

\State 
Input $H = H W$, $\gamma = 1 - 1/(HW)$, $\text{Data}$ and $\sigma, \lambda$ into Algorithm \ref{alg:stan_rlsvi_generic} and get output $\tilde{\theta}_{k}$. 
 
\For{Time block $w = 1, 2, \dots, W$}

\State Observe $s_{k,w}^{\high}$.

\For{Time period $h = 1, \dots, H$}
\State Observe $s_{k,w,h}^{\low}$.
\If{ $h = 1$}
\State Sample $\p{a_{k,w}^{\high}, a_{k,w, h}^{\low}}\in \operatorname{argmax}_{\alpha, \alpha'} \tilde{\theta}_{k} ^\top \phi \left(s^{\high}_{k,w}, \alpha, s^{\low}_{k,w, h}, \alpha' \right)$.
\Else
\State Sample $a_{k,w, h}^{\low} \in \operatorname{argmax}_{\alpha} \tilde{\theta}_{k} ^\top \phi \left(s^{\high}_{k,w}, a^{\high}_{k,w}, s^{\low}_{k,w, h}, \alpha \right)$.
\EndIf
\State Observe $r^{\low}_{k,w,h}$.
\State Add $(s_{k, (w-1)H + h}, a_{k, (w-1)H + h}, r_{k, (w-1)H + h})$ to Data with
\[
\begin{split}
s_{k, (w-1)H + h} &= \p{s_{k,w}^{\high}, a_{k,w}^{\high}, s_{k,w,h}^{\low}}, \qquad r_{k, (w-1)H + h} = r^{\low}_{k,w,h},\\
 a_{k, (w-1)H + h} &= 
  \begin{cases}
     \p{a_{k,w}^{\high}, a_{k,w,h}^{\low}} &\text{ if } h = 1,\\
     a_{k,w,h}^{\low} &\text{ otherwise}.
 \end{cases}
 \end{split}
 \]
\EndFor
\EndFor
\EndFor
\end{algorithmic}
\end{algorithm}


\begin{algorithm}
\caption{Bandit Algorithm (Episodic)}
\label{alg:bandit_baseline}
\begin{algorithmic}
\Require Feature mapping $\psi$; Parameters $\sigma_{\TS}>0, \lambda_{\TS}>0$. 
\State $\text{Data} = \cb{}$.
\For{Episode $k = 1, 2, \dots, K$} 
\For{Time block $w = 1, 2, \dots, W$}
\State Observe $s_{k,w}^{\high}$.
\For{Time period $h = 1, \dots, H$}
\State Observe $s_{k,w,h}^{\low}$.
\State 
Input $\text{Data}$ and $\sigma_{\TS}, \lambda_{\TS}$ into Algorithm \ref{alg:thompson_sampling_generic} and get output $\tilde{\theta}_{k,w,h}$.
\If{ $h = 1$}
\State Sample $\p{a_{k,w}^{\high}, a_{k,w, h}^{\low}}\in \operatorname{argmax}_{\alpha, \alpha'} \tilde{\theta}_{k,w,h} ^\top \psi \left(s^{\high}_{k,w}, \alpha, s^{\low}_{k,w, h}, \alpha' \right)$.
\Else
\State Sample $a_{k,w, h}^{\low} \in \operatorname{argmax}_{\alpha} \tilde{\theta}_{k,w,h} ^\top \psi \left(s^{\high}_{k,w}, a^{\high}_{k,w}, s^{\low}_{k,w, h}, \alpha \right)$.
\EndIf
\State Observe $r_{k,w,h}^{\low}$.
\State Add $(s, a, r)$ to Data with
\[
s = \p{s_{k,w}^{\high}, a_{k,w}^{\high},s_{k,w, h}^{\low}}, \qquad
a = \begin{cases}
     \p{a_{k,w}^{\high}, a_{k,w,h}^{\low}} &\text{ if } h = 1,\\
     a_{k,w,h}^{\low} &\text{ otherwise},
 \end{cases} \qquad
 r = r^{\low}_{k,w,h}.
 \]
\EndFor
\EndFor
\EndFor
\end{algorithmic}
\end{algorithm}

\end{document}